\documentclass[twocolumn]{article}

\usepackage[utf8]{inputenc}
\usepackage{graphicx}
\usepackage{amsmath}
\usepackage{booktabs}
\usepackage{bbm}
\usepackage{dashrule}
\usepackage{algorithm}
\usepackage{algorithmic}
\usepackage{color}
\usepackage{authblk}

\usepackage{url}
\usepackage{float}

\usepackage{eqparbox}

\usepackage{amsthm,amssymb}
\usepackage{booktabs}

\usepackage{tikz}
\usepackage{pgfplots}
\usepgfplotslibrary{fillbetween}
\usetikzlibrary{positioning}

\usepackage{comment}

\theoremstyle{definition}
\newtheorem{theorem}              {Theorem}

\newtheorem{proposition}[theorem] {Proposition}

\newtheorem{definition} [theorem] {Definition}

\newcommand{\x}{154}
\newcommand{\argmax}{\operatornamewithlimits{argmax}}

\title{Interpretable Companions for Black-Box Models}
\author[1]{Danqing Pan \thanks{danqing\_pan@ar.sanken.osaka-u.ac.jp}}
\author[2]{Tong Wang \thanks{tong-wang@uiowa.edu}}
\author[1]{Satoshi Hara \thanks{satohara@ar.sanken.osaka-u.ac.jp}}
\affil[1]{Osaka University, Japan}
\affil[2]{University of Iowa, USA}

\date{}

\begin{document}
\maketitle

\begin{abstract}
    We present an interpretable companion model for any pre-trained black-box classifiers. The idea is that for any input, a user can decide to either receive a prediction from the black-box model, with high accuracy but no explanations, or employ a companion rule to obtain an interpretable prediction with slightly lower accuracy. The companion model is trained from data and the predictions of the black-box model, with the objective combining area under the transparency--accuracy curve and model complexity. Our model provides flexible choices for practitioners who face the dilemma of choosing between always using interpretable models and always using black-box models for a predictive task, so users can, for any given input, take a step back to resort to an interpretable prediction if they find the predictive performance satisfying, or stick to the black-box model if the rules are unsatisfying. To show the value of companion models, we design a human evaluation on more than a hundred people to investigate the tolerable accuracy loss to gain interpretability for humans. 
\end{abstract}

%%%%%%%%%%%%%%%%%%%%%%%%%%%%%%%%%%%%%%%%%%%%%%%%%%%%%%%%%%%%%%%%%%%%%%%%
\section{Introduction}
The growing real-world needs for model understandability have triggered unprecedented advancement in the research in interpretable machine learning. Various forms of interpretable models have been created to compete with black-box models.  Given a predictive task, users need to choose between a black-box model with high accuracy but no interpretability and an interpretable model with compromised performance. However, in many practices, the end-users, instead of the model designers, need to have the flexibility of choosing between whether they need an interpretable prediction with the slightly compromised predictive performance or a non-interpretable prediction but higher task performance, based on the input case they have.

We design a new mechanism of making a prediction in this paper. For any input, we provide users two options, to use a black-box model with high accuracy or to use a \emph{companion} rule, which offers understandable prediction but with slightly lower accuracy.
It is up to the user which type of prediction s/he prefers for any input. The rules are embedded in an ``if-else'' logic structure with decreasing accuracy. Therefore, if a user goes deeper into the list, more companion rules will be activated to cover more instances, gaining higher model transparency, but more considerable performance loss is incurred. 

Our model is different from the current mainstream works in interpretable machine learning that focus on stand-alone interpretable models such as rule-based models \cite{wang2017bayesian} and linear models \cite{zeng2017interpretable}, or develops external black-box explanation methods \cite{ribeiro2016should,lundberg2017unified} to provide post-hoc analysis. The former models may suffer from possible loss in accuracy since, aside from optimizing predictive performance, they also need to optimize model interpretability in parallel, which often conflicts with model fitness to the data. The latter type of methods, on the other hand, undergo heated debate on whether they consistently and truthfully reflect the underlying synergies between features inside a black-box \cite{rudin2019stop}.

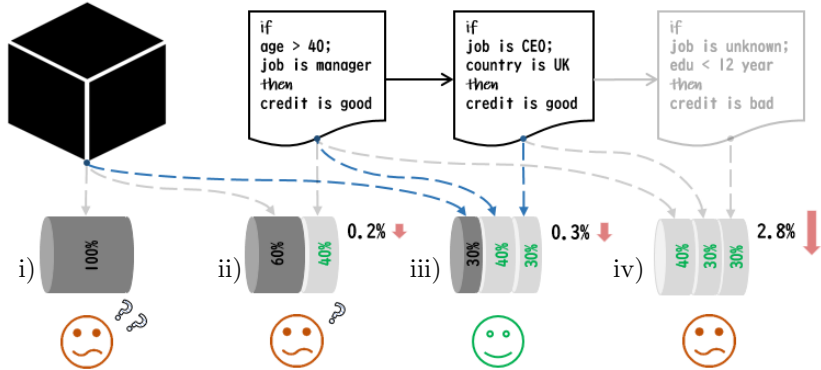
\begin{figure*}[t]
    \centering
    \begin{tikzpicture}
    \node[anchor=south west] at (0, 0){\includegraphics[scale=0.55]{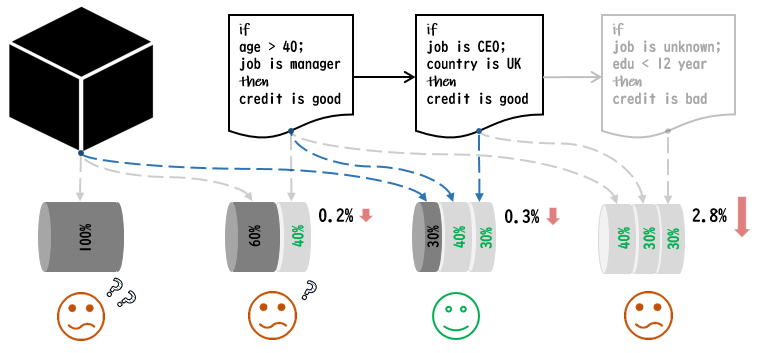}};
    \node at (0.5, 1.6){i)};
    \node at (3.2, 1.6){ii)};
    \node at (5.8, 1.6){iii)};
    \node at (8.5, 1.6){iv)};
    \end{tikzpicture}
    \vspace{-18pt}
    \caption{An example of evaluating customer credit with CRL. The users have four choices, i), ii), iii), and iv), for balancing the rule-based and the black-box predictions.}
    \label{fig:crl}
    \vspace{-8pt}
\end{figure*}

Our model, \emph{Companion Rule List} (CRL), takes a different route to avoid the possible weaknesses in the approaches above. Motivated by recent works on combining multiple models~\cite{wang2019hybrid,wang2015trading}, we pair a rule list with a pre-trained black-box model. Unlike previous hybrid models that create a fixed partition of the data space such that it is pre-determined which inputs will be processed by which model, we provide more flexibility to users by allowing users to switch between rules and the black-box, based on their task-specific and user-specific requirement for interpretability and accuracy.

\figurename~\ref{fig:crl} shows an example of CRL for evaluating customer credit.
In CRL, we assume a black-box model with superior predictive performance is already obtained.
CRL is designed to bring transparency into the decision making process as well as providing the freedom of selecting the rule-based and the black-box predictions.
In the figure, decision makers face four scenarios: i) using a completely black-box model; ii) adopting one rule that can explain 40\% of the inputs but with losing accuracy for 0.2\% and forwarding the rest 60\% to the black-box; iii) adopting two rules that can explain 70\% of the inputs but with losing accuracy for 0.3\% and forwarding the rest 30\% to the black-box; iv) adopting a completely transparent rule list but with losing accuracy for 2.8\%.
The final choice is made upon the users' preference on the trade-off between the transparency and the accuracy loss.
For example, one may prefer adopting the scenario iii) as a compromising solution to the trade-off.

Note that, the goal of CRL is not to train a stand-alone accurate rule list, but to train a rule list with a productive collaboration with the black-box model such that they achieve an efficient trade-off between interpretability and predictive performance. 
To train CRL with a superior trade-off, we propose a novel training algorithm that can take the knowledge of the black-box model into account.
Specifically, our algorithm is based on the area under the trade-off curve as the training objective function.

In the experiments, we demonstrate that the knowledge of the black-box partner during the training stage obtains a better performance for CRL, compared with other rule lists, such as CORELS \cite{angelino2017learning} and SBRL \cite{yang2017scalable} that are trained independently without knowledge of their black-box partner and paired with a black-box afterward.

To investigate whether users are willing to use companion rules instead of a black-box model despite possible performance loss, we conduct a carefully designed human evaluation on \x~participants with various background to study the conditions when a user would  switch to a companion rule, specifically, how much accuracy a user is willing to give up in exchange for interpretability. 

%The rest of the paper is organized as follows.
%Section \ref{sec:related} reviews related work in machine learning model interpretation.
%Section \ref{sec:companion} presents our proposed framework, CRL, for the collaboration of rule list and black-box.
%In Section \ref{sec:objective}, we propose a novel learning framework for CRL, and present a CRL training algorithm in Section \ref{sec:algorithm}.
%Section \ref{sec:experiment} provides an evaluation of CRL on data sets in several different domains, as well as a survey on human subjects.

%%%%%%%%%%%%%%%%%%%%%%%%%%%%%%%%%%%%%%%%%%%%%%%%%%%%%%%%%%%%%%%%%%%%%%%%
%\vspace{-2mm}
\subsection{Preliminaries}
%\vspace{-2mm}
\paragraph{Notation}
For any statement $a$, we denote the indicator function by $\mathbbm{1}(a)$ where $\mathbbm{1}(a) = 1$ if the statement $a$ is true, and $\mathbbm{1}(a) = 0$ otherwise.

%\vspace{-2mm}
\paragraph{Rule List}
Let $(x, y) \in \mathcal{X} \times \{0, 1\}$ be an observation consisting of an input feature vector $x$ in a certain domain $\mathcal{X}$ and the output class label $y$.
The rule list describes its prediction process in a "if-else-" format.
Let the \emph{decision function} $d$ be a map from $\mathcal{X}$ to the Boolean domain, i.e.\ $d: \mathcal{X} \to \{\mathrm{True}, \mathrm{False}\}$.
We refer to the pair of a decision function $d$ and an output $z \in \{0, 1\}$ as a \emph{rule} $r := (d, z)$.
Here, we read the rule $r$ as ``if $d(x) = \mathrm{True}$ for an input $x$, then the output $\hat{y} = z$''. $d(x) = \mathrm{True}$ means $x$ satisfies the conditions in the rule.
The rule list of length $M$ consists of the sequence of $M$ rules $R := (r_m = (d_m, z_m))_{m=1}^M$.
The rule list returns the prediction $\hat{y}$ based on the following format.
\begin{center}
    \begin{tabular}{l}
    \textbf{if} $d_1(x) = \mathrm{True}$, \textbf{then} $\hat{y}=z_1$\\
    \textbf{else if} $d_2(x) = \mathrm{True}$, \textbf{then} $\hat{y}=z_2$\\
    $\ldots$\\
    \textbf{else if} $d_{M}(x) = \mathrm{True}$, \textbf{then} $\hat{y}=z_{M}$
    \end{tabular}
\end{center}
We denote the prediction of the rule list by $\hat{y} = \mathcal{R}_M(x ; R)$.
%Below we define the cover and prediction of rule lists.
We also define the cover of the rule list as follows.
\begin{definition}[Cover]
    We say that the $m$-th rule $r_m = (d_m, z_m)$ covers the input $x$ if $d_m(x) = \mathrm{True}$ and $d_k(x) = \mathrm{False}$ for any of the previous rule $r_k = (d_k, z_k)$ with $k < m$. 
    For the rule $r_m$ and the input $x$, we define $\mathrm{covers}(r_m, x)$ by
    \begin{align}
        \mathrm{covers}(r_m, x) := \mathbbm{1}\left(\left(\bigwedge_{k=1}^{m-1}\neg d_k(x)\right) \wedge d_m(x)\right).
    \end{align}
    Note that $\mathrm{covers}(r_m, x) = 1$ if the input $x$ is covered by $r_m$, and $\mathrm{covers}(r_m, x) = 0$ otherwise. Thus, each input $x$ is assigned to the first rule it satisfies.
\end{definition}
%\begin{definition}[Rule List Prediction]
%    The prediction of the rule list of length $M$, consisting of the sequence of $M$ rules $R := (r_m = (d_m, z_m))_{m=1}^M$, is defined by
%    \begin{align}
%        \hat{y} = \mathcal{R}_M(x ; R) := \sum_{m=1}^M z_m\mathrm{covers}(r_m, x).
%    \end{align}
%\end{definition}

%%%%%%%%%%%%%%%%%%%%%%%%%%%%%%%%%%%%%%%%%%%%%%%%%%%%%%%%%%%%%%%%%%%%%%%%
\section{Related Work}
\label{sec:related}
%In this section, we first elaborate distinctions from current interpretable machine learning researches and then discuss the similarity with techniques we inherit from existing works.
We elaborate distinctions from current interpretable machine learning researches and discuss the similarity with techniques we inherit from existing works.

%\vspace{-2mm}
\paragraph{Distinction from Black-box Explainers}
CRL is not a black-box explainer.
The rules are not designed to explain or approximate black-boxes like the rule-based explainers do \cite{ribeiro2018anchors,guidotti2018local}.
CRL is constructed to \emph{compete with} and \emph{locally replace} a black-box with competitive performance. 
We emphasize this distinction because black-box explainers serve the purpose of providing post-hoc analysis and approximations to the decision maker, which may not represent the true interactions of features inside the black-box \cite{rudin2019stop,aivodji2019fairwashing}.
CRL, on the other hand, is the decision maker itself, and thus it truly represents how the decision is made.

%\vspace{-2mm}
\paragraph{Rule-Based Models}
Our CRL consists of decision rules.
Rules are a well-adopted form of interpretable models for their language-like presentation and simple logic.
Many state-of-the-art interpretable models are rule sets \cite{wang2017bayesian,wang2018multi,dash2018boolean,lakkaraju2016interpretable} or rule lists \cite{angelino2017learning,yang2017scalable,aivodji2019learning} constructed via various learning algorithms like simulated annealing \cite{kirkpatrick1983optimization}.
%The idea is to propose a new model by making small changes to the current model, accept it with decreasing probability which is a function of how good the proposal is and the current temperature, until the maximum iterations are met or the solution converges.
%The idea is to propose a new model by making small changes to the current model, accept it with decreasing probability which is a function of how good the proposal is and the current temperature, until the maximum iterations are met or the solution converges.
The idea is to propose a new model by making small changes to the current model and accept it a with decreasing probability until the maximum iterations are met or the solution converges.
We design our training algorithm builds upon the prior wisdom from the works above and incorporate new strategies exploiting the unique structure of CRL.

%\vspace{-2mm}
\paragraph{Hybrid Models}
CRL is one type of companion models. Wang et al.~\cite{wang2015trading} proposed to divide feature spaces into regions with sparse oblique trees and then assign black-box local experts to each region.
Nan and Saligrama~\cite{nan2017adaptive} designed a low-cost adaptive system by training a gating and prediction model that limits the utilization of a high-cost model to hard input instances and gates easy-to-handle input instances to a low-cost model.
The work closest to ours is hybrid models \cite{wang2019hybrid,wang2019gaining} that partition the feature space into transparency areas that are covered by rules and black-box area where rules fail to characterize.

Compared with our CRL, the models mentioned above create a fixed partition when the model is built.
Thus it is determined at the training stage, by the model designer, which predictions will be processed by which model, and what level of transparency will be provided.
CRL, on the other hand, leaves this decision to the end-user of the model, who may switch to different transparency and accuracy pairs for different inputs, accounting for various task-specific complications. Such flexibility is very critical and practical in domains where the final decision maker is human.

%%%%%%%%%%%%%%%%%%%%%%%%%%%%%%%%%%%%%%%%%%%%%%%%%%%%%%%%%%%%%%%%%%%%%%%%
\section{Companion Rule List}
\label{sec:companion}

We propose \emph{Companion Rule List} (CRL) as a collaboration framework for the interpretable and the black-box models.
The advantage of our CRL lies in its flexibility that the users can choose which model to adopt for any inputs.

%%%%%%%%%%%%%%%%%%%%%%%%%%%%%%%%%%%%%%%%%%%%%%%%%%%%%%%%%%%%%%%%%%%%%%%%
\subsection{The Proposed Framework}
Let $f_b$ be a pre-obtained black-box model, such as the ensemble models or deep neural networks, that have superior predictive performance.
CRL is a collaboration framework for the rule list and the black-box model.
CRL is expressed in the following form.
\begin{center}
    \begin{tabular}{l}
    \textbf{if} $d_1(x) = \text{True}$, \textbf{then} $\hat{y}=z_1$ \textbf{or} $\hat{y}=f_b(x)$\\
    \textbf{else if} $d_2(x) = \text{True}$, \textbf{then} $\hat{y}=z_2$ \textbf{or} $\hat{y}=f_b(x)$\\
    $\ldots$\\
    \textbf{else if} $d_{M}(x) = \text{True}$, \textbf{then} $\hat{y}=z_{M}$ \textbf{or} $\hat{y}=f_b(x)$\\
    \end{tabular}
\end{center}

Note that CRL is flexible by its design. 
For any input $x$, if $x$ is covered by the $m$-th rule as $d_m(x) = \mathrm{True}$, the users have a choice of adopting the output $\hat{y}=z_m$ from the rule list or the output $\hat{y}=f_b(x)$ from the black-box model.

As we pointed out in Section~\ref{sec:related}, the hybrid rule set (HRS)~\cite{wang2019hybrid,wang2019gaining} also provides a framework for the collaboration of the interpretable and the black-box models.
Despite the similarity, we would like to emphasize that HRS is not flexible.
The rule set in the HRS covers a pre-determined fraction of the inputs. The end-users cannot choose between rules and the black-box model based on their own needs.

%%%%%%%%%%%%%%%%%%%%%%%%%%%%%%%%%%%%%%%%%%%%%%%%%%%%%%%%%%%%%%%%%%%%%%%%
\subsection{A Naive Implementation}\label{sec:naive}
A straightforward way to implement a companion model is to combine an independently trained rule list with the black-box model.
For example, one can train a rule list by using CORELS \cite{angelino2017learning} and SBRL \cite{yang2017scalable}.
Then, simply combining the trained rule list with the black-box model yields a companion model.
This approach is straightforward, but the result can be suboptimal, as will be demonstrated by experiments. This is because the collaboration with the black-box model is not taken into account when training the rule list.

%%%%%%%%%%%%%%%%%%%%%%%%%%%%%%%%%%%%%%%%%%%%%%%%%%%%%%%%%%%%%%%%%%%%%%%%
\section{The Learning Framework for CRL}
\label{sec:objective}

We discuss how we will construct a CRL to better collaborate with a black-box. To do that, we propose a novel objective function  so that the interpretable and the black-box models collaborate more effectively than the naive implementation.

In what follows, we assume that an observation $(x, y) \in \mathcal{X} \times \{0, 1\}$ is sampled independently from an underlying distribution $p$.
We also denote a set of independent observations $(x, y)$ from $p$ as the data set $D$, and denote its size by $|D|$.

%%%%%%%%%%%%%%%%%%%%%%%%%%%%%%%%%%%%%%%%%%%%%%%%%%%%%%%%%%%%%%%%%%%%%%%%
\subsection{Area Under the Transparency--Accuracy Trade-off Curve (AUTAC)}

We propose \emph{the area under the transparency--accuracy trade-off curve} (AUTAC) as a metric for general companion models.

Recall that we expect a companion model to be flexible so that the users can freely switch between the interpretable and the black-box models.
Suppose that a user is willing to use the interpretable model for 100$t$\% of the observations (with $0 \le t \le 1$), and the black-box model for the remaining observations.
Here, we refer to $t$ as \emph{transparency} of the companion model, which is a fraction of observations explained by the interpretable model.
Note that a user preferring small $t$ favors the black-box model because of its higher accuracy, while a user preferring large $t$ favors the interpretable model.
The difficulty here is that such a user's preference on $t$ is unknown in practice.
To bypass this difficulty, we adopt the following principle.

%\vspace{-2mm}
\textbf{Maximum Accuracy Principle}\;
For any transparency $t$, a user prefers high accuracy.

From this principle, an ideal companion model is one that maximizes the accuracy for all transparency $t$.
To this end, we propose the area under the transparency--accuracy trade-off curve (AUTAC) as the goodness measure of a companion model.
Formally, let $c_t$ be a companion model whose transparency is $t$, and let $A(c_t) := \mathbb{E}_{(x, y) \sim p}[\mathbbm{1}(y = c_t(x)]$ be an accuracy of $c_t$.  
We can then draw a curve $(t, A(c_t))$ representing the transparency--accuracy trade-off.
The area under the curve, or AUTAC, is then defined by
\begin{align}
    \mathrm{AUTAC} = \int_{t=0}^1 A(c_t) dt.
    \label{eq:autac}
\end{align}

%%%%%%%%%%%%%%%%%%%%%%%%%%%%%%%%%%%%%%%%%%%%%%%%%%%%%%%%%%%%%%%%%%%%%%%%
\subsection{AUTAC of CRL}
We first define some important notions, and then derive AUTAC of CRL.
Note that we generate rule lists with decreasing accuracy of rules.
Thus, the maximum accuracy principle above implies that the users always adopt the first $m$ consecutive rules in CRL as the interpretable model, where $m$ can vary depending on their preferences on $t$.

\begin{definition}[Transparency of Rule List]
    For a rule list $\mathcal{R}_M(R)$ with a sequence of rules $R = (r_m = (d_m, z_m))_{m=1}^M$, the transparency of the first $m$ rules is the probability of observations covered by these rules, defined by
    \begin{align}
        T_m := \mathbb{E}_{(x, y) \sim p}\left[ \mathbbm{1}\left(\bigvee_{k=1}^m \mathrm{covers}(r_k, x)\right) \right] .
    \end{align}
\end{definition}
Note that $T_m$ takes a value between zero and one.
The value one indicates that all observations can be explained by the first $m$ rules, and thus its decision process is completely transparent.
The value zero indicates that none of the observations are covered by the first $m$ rules.

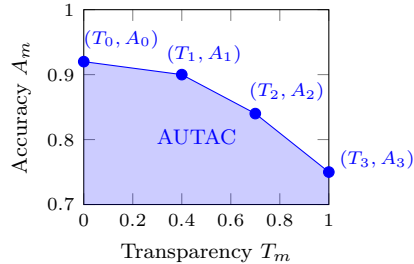
\begin{figure}[t]
    \centering
    \begin{tikzpicture}
    \begin{axis}[
        width=0.6\linewidth,
        xmin=0, xmax=1,
        ymin=0.7, ymax=1,
        xlabel={\footnotesize Transparency $T_m$},
        ylabel={\footnotesize Accuracy $A_m$},
        tick label style={font=\scriptsize},
        x label style={at={(axis description cs:0.5,0.05)},anchor=north},
        y label style={at={(axis description cs:0.2,0.5)},anchor=south},
    ]
    \addplot[name path=tradeoff, blue,mark=*] coordinates{(0,0.92)(0.4,0.90)(0.7,0.84)(1.0,0.75)};
    \path[name path=axis] (axis cs:0,0) -- (axis cs:1,0);
    \addplot [fill=blue,opacity=0.2,forget plot] fill between[of=tradeoff and axis];
    \end{axis}
    \node at (0.5, 2.2){\scriptsize \textcolor{blue}{$(T_0, A_0)$}};
    \node at (1.6, 2.0){\scriptsize \textcolor{blue}{$(T_1, A_1)$}};
    \node at (2.7, 1.5){\scriptsize \textcolor{blue}{$(T_2, A_2)$}};
    \node at (3.9, 0.6){\scriptsize \textcolor{blue}{$(T_3, A_3)$}};
    \node at (1.5, 0.9){\footnotesize \textcolor{blue}{AUTAC}};
    \end{tikzpicture}
    \caption{An example of the transparency--accuracy trade-off curve and the AUTAC of a Stochastic CRL with length $M = 3$.}
    \label{fig:tradeoff}
\end{figure}

\begin{definition}[Stochastic CRL]
    Let $m^t := \max_{T_m \le t} m$ be the maximum number of rules whose transparency is at most $t$.
    We define a Stochastic CRL with a transparency $t$ by
    
    \medskip
        \begin{tabular}{l}
        \textbf{if} $\bigvee_{k=1}^{m^t} \mathrm{covers}(r_k, x) = \mathrm{True}$, \textbf{then} $y=\mathcal{R}_{M}(x ; R)$\\
        \textbf{else if} $\bigvee_{k=1}^{m^t+1} \mathrm{covers}(r_k, x) = \mathrm{False}$, \textbf{then} $y=f_b(x)$\\
        \textbf{else} $y=\begin{cases}z_{m^t} & \text{if $q_t > \varepsilon$} \\  f_b(x) & \text{otherwise} \end{cases}$
        \end{tabular}
        
    \medskip
    where $q_t := \frac{t - T_{m^t}}{T_{m^t+1} - T_{m^t}} \in [0, 1]$, and $\varepsilon$ is a uniform random variable in $[0, 1]$.
\end{definition}
\normalsize
In Stochastic CRL, the users stochastically determine which model to adopt based on the transparency $t$ and the random variable $\varepsilon$.
Note that, the expected transparency of the Stochastic CRL with respect to $\varepsilon$ can be easily verified as 
$q_t T_{m^t+1} + (1 - q_t) T_{m^t} = \frac{t - T_{m^t}}{T_{m^t+1} - T_{m^t}} T_{m^t+1} + \frac{T_{m^t+1} - t}{T_{m^t+1} - T_{m^t}} T_{m^t} = t$.

%%%%%%%%%%%%%%%%%%%%%%%%%%%%%%%%%%%%%%%%%%%%%%%%%%%%%%%%%%%%%%%%%%%%%%%%
%\vspace{-2mm}
\paragraph{Defining AUTAC for CRL}

We define the AUTAC of CRL by adopting Stochastic CRL as the companion model $c_t$.
Here, we use Stochastic CRL because its expected transparency $t$ is continuous and the integral (\ref{eq:autac}) is well-defined, while the transparency of the original CRL is defined only on discrete points $T_1, T_2, \ldots, T_M$.
Let the accuracy of the Stochastic CRL be $A_m := \mathbb{E}_{\varepsilon}[A(c_{T_m})]$.
We can then draw the transparency--accuracy trade-off curve as shown in \figurename~\ref{fig:tradeoff}.
Moreover, it is clear from the figure that AUTAC can be computed as follows.
\begin{proposition}[AUTAC of Stochastic CRL]
    The AUTAC (\ref{eq:autac}) with Stochastic CRL as $c_t$ is given by
    %\begin{align}
    %    \mathrm{AUTAC}_M(R) = \frac{1}{2} \sum_{m=1}^M (A_m + A_{m-1})(T_m - T_{m-1}) .
    %    \label{eq:auc}
    %\end{align}
    \begin{align*}
        \mathrm{AUTAC}_M(R) = \frac{1}{2} \sum_{m=1}^M (A_m + A_{m-1})(T_m - T_{m-1}) .
    \end{align*}
    Here, we expressed the dependency of AUTAC to the sequence of the rules $R$ and its length $M$, explicitly.
\end{proposition}

%%%%%%%%%%%%%%%%%%%%%%%%%%%%%%%%%%%%%%%%%%%%%%%%%%%%%%%%%%%%%%%%%%%%%%%%
%\vspace{-2mm}
\subsection{Learning Objective}

We propose to train CRL so that the AUTAC is maximized.
As a training objective, we use the following estimation of the AUTAC.
\begin{align*}
    \widehat{\mathrm{AUTAC}}_M(R) = \frac{1}{2} \sum_{m=1}^M (\hat{A}_m + \hat{A}_{m-1})(\hat{T}_m - \hat{T}_{m-1}),
\end{align*}
where 
\begin{align*}
    \hat{T}_m :=&\, \textstyle{\frac{|S_m|}{|D|}},  \\
    \hat{A}_m :=&\, \textstyle{\frac{1}{|D|} \sum_{(x, y) \in S_m} \mathbbm{1}[y=\mathcal{R}_m(x; R)]} \\
    &+ \textstyle{\frac{1}{|D|} \sum_{(x, y) \in D \setminus S_m} \mathbbm{1}[y = f_b(x)]}, \\
    S_m :=&\, \textstyle{\left\{(x, y) \,\middle\vert\, \bigvee_{k=1}^m \mathrm{covers}(r_k, x) = \mathrm{True}, (x, y) \in D \right\}}.\\
\end{align*}

The training of CRL is then formulated as
\begin{align}
    \max_M \max_{R} O_{M, \alpha}(R) := \widehat{\mathrm{AUTAC}}_M(R) - \alpha M,
    \label{eq:obj}
\end{align}
where $\alpha \ge 0$ is a parameter that penalizes the length of the rule list.
Here, the additional term $\alpha M$ enforces the rule list to be sufficiently short so that it exhibits good interpretability.

%%%%%%%%%%%%%%%%%%%%%%%%%%%%%%%%%%%%%%%%%%%%%%%%%%%%%%%%%%%%%%%%%%%%%%%%
\section{Training Algorithm}
\label{sec:algorithm}

We now design a training algorithm for CRL.
Note that the problem (\ref{eq:obj}) is a combinatorial optimization problem and finding a global optimum can take exponential time in general.
We therefore take an alternative approach based on the heuristc search.
Specifically, we adopt the stochastic local search for our training algorithm, which is shown to be effective for training high-quality rule models~\cite{lakkaraju2016interpretable,wang2019gaining}.

The proposed training algorithm is shown in Algorithm~\ref{alg:A}.
At the initilization step, we first apply FP-Growth~\cite{han2000mining} to the data set $D$, and find both the frequent positive rules and the frequent negative rules with the minimum support bounded by $\gamma$, and construct the set of rules $\Gamma$.\footnote{
Other rule miners such as Apriori or Eclat can also be used instead of FP-Growth.}
We then initialize the rule list with three rules chosen (possible randomly) from $\Gamma$.
After the initialization is completed, we iteratively update the model using the stochastic local search.
There are four possible operations for the model update: add, remove, swap, and replace.
\vspace{-0.8\baselineskip}
\begin{itemize}
    \setlength{\itemsep}{2pt}
    \setlength{\parskip}{2pt}
    \item \textbf{Add}: We randomly select one rule from $\Gamma$. We then insert the selected rule into a random position in the current rule list.
    \item \textbf{Remove}: We randomly select one rule from the current rule list, and remove it from the list.
    \item \textbf{Switch}: We randomly select two rules in the current list, and swap their positions.
    \item \textbf{Replace}: We randomly select one rule from $\Gamma$. We also select one rule from the current rule list. We then replace these two rules.
\end{itemize}
\vspace{-0.8\baselineskip}
In each update step, one of the four operations is randomly chosen and applied to the rule list.
The model update is accepted with probability $\exp\left(\frac{O_{M,\alpha}(R^{[n+1]})-O_{M,\alpha}(R^{[n]})}{C_{0}/\log_{2}(1+n)}\right)$ which gradually decreases as the iteration $n$ increases due to annealing.

\begin{algorithm}[t]
    \caption{Stochastic Local Search for CRL}
    \label{alg:A} 
    \begin{algorithmic}
        \STATE {\textbf{Input:} $f_{b},D,\alpha,C_{0}$} 
        \vspace{3pt}
        \STATE{$\triangleright$\texttt{Initialization}}
        \STATE {$\Gamma = ((d_j, z_j))_{j=1}^J \leftarrow \mathrm{FPGrowth}(D, \mathrm{minsupp} = \gamma) $\\$ \hspace{8pt} \triangleright$ mine candidate rules from $D$} 
        \STATE {$R^{[0]} \leftarrow ((d_j, z_j))_{j=1}^3, R^* \leftarrow ((d_j, z_j))_{j=1}^3, M \leftarrow 3$ \\$ \hspace{8pt} \triangleright$ initialize the rule list and the best solution}
        \\\hdashrule{8cm}{1pt}{1pt}
        %\vspace{5pt}
        \STATE{$\triangleright$\texttt{Stochastic Local Search}}
        \FOR{$n=1, 2, \ldots, N$}
            \STATE {$\delta \sim \mathrm{random}()$}
            \IF[\hspace{6pt} $\triangleright$\texttt{Add}]{$\delta < \frac{1}{4}$}
                \STATE {$R^{[n+1]} \leftarrow$ add one rule to $R^{[n]}$ from $\Gamma$}
                \STATE {$M \leftarrow M+1$}
            \ELSIF[\hspace{6pt} $\triangleright$\texttt{Remove} 
            ]{$\delta < \frac{1}{2}$}
                \STATE {$R^{[n+1]} \leftarrow$ remove one rule from $R^{[n]}$}
                \STATE {$M \leftarrow M-1$}
            \ELSIF[\hspace{6pt} $\triangleright$\texttt{Swap}]{$\delta < \frac{3}{4}$}
                \STATE {$R^{[n+1]} \leftarrow$ swap two rules in $R^{[n]}$}
            \ELSE[\hspace{6pt} $\triangleright$\texttt{Replace}]
                \STATE {$R^{[n+1]} \leftarrow$ replace a rule in $R^{[n]}$ with a rule in $\Gamma$}
            \ENDIF
            \vspace{5pt}
            \STATE{$\triangleright$\texttt{Model Update}}
            \STATE {$\epsilon \sim \mathrm{random}()$}
            \IF{$\epsilon > \exp\left(\frac{O_{M, \alpha}(R^{[n+1]})-O_{M, \alpha}(R^{[n]})}{C_{0}/\log_{2}(1+n)}\right)$}
                \STATE {$R^{[n+1]} \leftarrow R^{[n]}$}
            \ENDIF
            \STATE {$R^{*} \leftarrow \argmax_{R \in \{R^{[n+1]},R^{*}\}} O_{M, \alpha}(R)$ \\ \hspace{8pt} $\triangleright$ update the best model}
        \ENDFOR
        \STATE {\textbf{output $\mathcal{R}_M(x; R^{*})$} \hspace{6pt} $\triangleright$\texttt{Trained Rule List}}
\end{algorithmic}
\end{algorithm}

%%%%%%%%%%%%%%%%%%%%%%%%%%%%%%%%%%%%%%%%%%%%%%%%%%%%%%%%%%%%%%%%%%%%%%%%
\section{Experiments}
\label{sec:experiment}

In the first part of this section, we test the performance of  CRL  on public datasets and compare it with two independently trained rule lists CORELS~\cite{angelino2017learning} and SBRL~\cite{yang2017scalable}.\footnote{An example code is available at \url{https://github.com/danqingpan/Companion-rule-list}}
In the second part, a human evaluation of transparency--accuracy trade-off is conducted.

%%%%%%%%%%%%%%%%%%%%%%%%%%%%%%%%%%%%%%%%%%%%%%%%%%%%%%%%%%%%%%%%%%%%%%%%
%\vspace{-2mm}
\subsection{Experiments on Public Datasets}

We evaluate the transparency--accuracy trade-off of CRL on eight public data sets from UCI machine learning repository \cite{Dua:2019} or ICPSR, listed in \tablename~\ref{tab:datasets}.
Six data sets are associated with medical, financial, and judicial areas that have intensive demands for interpretable models.
The other two data sets are associated with physics and commerce.

\vspace{-4mm}
\paragraph{Preprocessing}
We preprocess the data by turning raw input into binary features so that rule mining is applicable.
To do that, we convert numerical features to categorical features by using the quantile-based quantization.\footnote{We use \texttt{quantile}\_\texttt{transform} in \texttt{scikit-learn} with the number of quantiles set to seven.}
We then apply one-hot encoding for categorical features (including the ones converted from the numerical features) to obtain binary features.

\vspace{-4mm}
\paragraph{Baselines}
As discussed in section \ref{sec:naive}, we adopt the two rule list training algorithms, CORELS\footnote{\url{https://github.com/fingoldin/pycorels}}~\cite{angelino2017learning} and SBRL\footnote{\url{https://github.com/Hongyuy/sbrlmod}}~\cite{yang2017scalable} as the baseline methods to be compared with.
Note that, these algorithms does not utilize the information of their black-box partner when training the rule list.
Thus, the obtained models from these training algorithms will become suboptimal, and will be outperformed by our proposed training algorithm.
We also apply decision trees, C4.5~\cite{quinlan2014c4} and CART~\cite{breiman2017classification}, to work as stand-alone interpretable baselines because they are one of the most popular interpretable models.

\begin{table}[t]
    \vspace{-12pt}
    \centering
    \small
    \caption{Data Sets: $N$, $d$, and $d'$ denote the number of instance, the number of features, and the number of binary features after preprocessing, respectively.}
    \label{tab:datasets}
    \begin{tabular}{llcccc}
        \toprule
       \textbf{datasets} &\textbf{category} & \textbf{$N$} & \textbf{$d$} & \textbf{$d'$} \\
        \midrule
        messidor & medical & 1151 & 19 & 202 \\
        german & finance & 1000 & 20 & 160 \\
        adult & finance & 48842 & 14 & 126 \\
        juvenile & justice & 4023 & 55 & 315\\
        frisk & justice & 80755 & 26 & 92\\
        recidivism & justice & 11645 & 106 & 641 \\
        magic & physics & 19020 & 10 & 140 \\
        coupon & commerce & 3996 & 95 & 95\\
        \bottomrule
    \end{tabular}
    \vspace{-12pt}
\end{table}
\normalsize

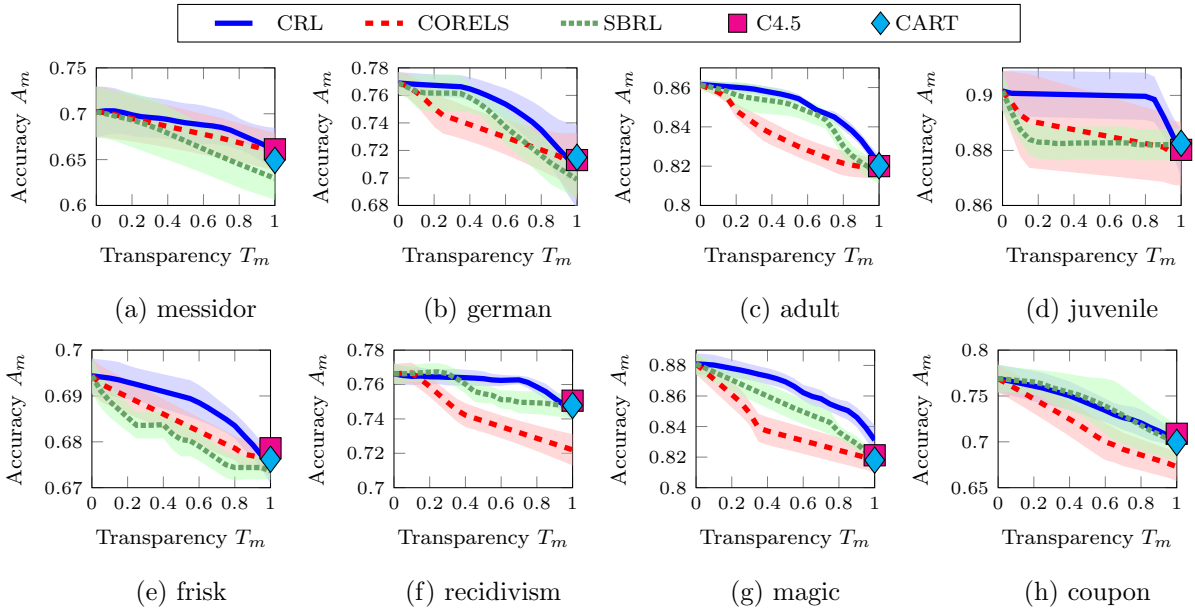
\begin{figure*}[h]
    \centering
    \begin{tikzpicture}
    \node[anchor=west] at (0,0) [draw,minimum width=11.2cm,minimum height=0.5cm] {};
    \node[anchor=west](crl) at (1.2, 0) {\footnotesize CRL};
    \draw[blue, solid, line width=2pt] (0.5, 0) -- (1.0, 0);
    \node[anchor=west, right=1.0 of crl](corels) {\footnotesize CORELS};
    \draw[red, dashed, line width=2pt] (2.5, 0) -- (3.0, 0);
    \node[anchor=west, right=1.0 of corels](sbrl) {\footnotesize SBRL};
    \draw[green!40!black!60!, densely dotted, line width=2pt] (5.1, 0) -- (5.6, 0);
    \node[anchor=west, right=1.0 of sbrl](c45) {\footnotesize C4.5};
    \draw[black,fill=magenta!99!white] (7.57,0.12) -- (7.57,-0.12) -- (7.33,-0.12) -- (7.33,0.12) -- cycle;
    \node[anchor=west, right=1.0 of c45](cart) {\footnotesize CART};
    \draw[black,fill=cyan!99!white] (9.35,0.15) -- (9.47,0.0) -- (9.35,-0.15) -- (9.23,0.0) -- cycle;
    \end{tikzpicture}\\
    \vspace{3pt}
  \def\file{./figs/messidor_rf_avg.csv}%
  \def\tree{./figs/messidor_tree.csv}%
  \def\ymax{0.75}%
  \def\ymin{0.60}%
  \def\leg{(a) messidor}%
  \begin{tikzpicture}
\begin{axis}[
    name=axis,
    width=0.24\textwidth,
    xmin=0, xmax=1,
    ymin=\ymin, ymax=\ymax,
    xlabel={\footnotesize Transparency $T_m$},
    ylabel={\footnotesize Accuracy $A_m$},
    tick label style={font=\scriptsize},
    x label style={at={(axis description cs:0.5,0.05)},anchor=north},
    y label style={at={(axis description cs:0.2,0.5)},anchor=south},
]
\addplot[name path=companion_upper,draw=none,forget plot] table[x=cover,y expr=\thisrow{companion_acc_avg}+\thisrow{companion_acc_std}, col sep=comma] {\file};
\addplot[name path=companion_lower,draw=none,forget plot] table[x=cover,y expr=\thisrow{companion_acc_avg}-\thisrow{companion_acc_std}, col sep=comma] {\file};
\addplot [fill=blue!30,opacity=0.5,forget plot] fill between[of=companion_upper and companion_lower];

\addplot[name path=corels_upper,draw=none,forget plot] table[x=cover,y expr=\thisrow{corels_acc_avg}+\thisrow{corels_acc_std}, col sep=comma] {\file};
\addplot[name path=corels_lower,draw=none,forget plot] table[x=cover,y expr=\thisrow{corels_acc_avg}-\thisrow{corels_acc_std}, col sep=comma] {\file};
\addplot [fill=red!30,opacity=0.5,forget plot] fill between[of=corels_upper and corels_lower];

\addplot[name path=sbrl_upper,draw=none,forget plot] table[x=cover,y expr=\thisrow{sbrl_acc_avg}+\thisrow{sbrl_acc_std}, col sep=comma] {\file};
\addplot[name path=sbrl_lower,draw=none,forget plot] table[x=cover,y expr=\thisrow{sbrl_acc_avg}-\thisrow{sbrl_acc_std}, col sep=comma] {\file};
\addplot [fill=green!30,opacity=0.5,forget plot] fill between[of=sbrl_upper and sbrl_lower];

\addplot[blue, solid, line width=2pt] table[x=cover,y=companion_acc_avg,col sep=comma] {\file};
\addplot[red, dashed, line width=2pt] table[x=cover,y=corels_acc_avg,col sep=comma] {\file};
\addplot[green!40!black!60!, densely dotted, line width=2pt] table[x=cover,y=sbrl_acc_avg,col sep=comma] {\file};

\addplot[black, fill=magenta!99!white, mark=square*, mark size=4pt] table[x=cover,y=c45_acc_avg,col sep=comma] {\tree};
\addplot[black, fill=cyan!99!white, mark=diamond*, mark size=5pt] table[x=cover,y=cart_acc_avg,col sep=comma] {\tree};

\end{axis}
\node[below=1.1cm of axis.south] {\leg};

\end{tikzpicture}%

    \hspace{-4pt}
  \def\file{./figs/german_rf_avg.csv}%
  \def\tree{./figs/german_tree.csv}%
  \def\ymax{0.78}%
  \def\ymin{0.68}%
  \def\leg{(b) german}%
  \begin{tikzpicture}
\begin{axis}[
    name=axis,
    width=0.24\textwidth,
    xmin=0, xmax=1,
    ymin=\ymin, ymax=\ymax,
    xlabel={\footnotesize Transparency $T_m$},
    ylabel={\footnotesize Accuracy $A_m$},
    tick label style={font=\scriptsize},
    x label style={at={(axis description cs:0.5,0.05)},anchor=north},
    y label style={at={(axis description cs:0.2,0.5)},anchor=south},
]
\addplot[name path=companion_upper,draw=none,forget plot] table[x=cover,y expr=\thisrow{companion_acc_avg}+\thisrow{companion_acc_std}, col sep=comma] {\file};
\addplot[name path=companion_lower,draw=none,forget plot] table[x=cover,y expr=\thisrow{companion_acc_avg}-\thisrow{companion_acc_std}, col sep=comma] {\file};
\addplot [fill=blue!30,opacity=0.5,forget plot] fill between[of=companion_upper and companion_lower];

\addplot[name path=corels_upper,draw=none,forget plot] table[x=cover,y expr=\thisrow{corels_acc_avg}+\thisrow{corels_acc_std}, col sep=comma] {\file};
\addplot[name path=corels_lower,draw=none,forget plot] table[x=cover,y expr=\thisrow{corels_acc_avg}-\thisrow{corels_acc_std}, col sep=comma] {\file};
\addplot [fill=red!30,opacity=0.5,forget plot] fill between[of=corels_upper and corels_lower];

\addplot[name path=sbrl_upper,draw=none,forget plot] table[x=cover,y expr=\thisrow{sbrl_acc_avg}+\thisrow{sbrl_acc_std}, col sep=comma] {\file};
\addplot[name path=sbrl_lower,draw=none,forget plot] table[x=cover,y expr=\thisrow{sbrl_acc_avg}-\thisrow{sbrl_acc_std}, col sep=comma] {\file};
\addplot [fill=green!30,opacity=0.5,forget plot] fill between[of=sbrl_upper and sbrl_lower];

\addplot[blue, solid, line width=2pt] table[x=cover,y=companion_acc_avg,col sep=comma] {\file};
\addplot[red, dashed, line width=2pt] table[x=cover,y=corels_acc_avg,col sep=comma] {\file};
\addplot[green!40!black!60!, densely dotted, line width=2pt] table[x=cover,y=sbrl_acc_avg,col sep=comma] {\file};

\addplot[black, fill=magenta!99!white, mark=square*, mark size=4pt] table[x=cover,y=c45_acc_avg,col sep=comma] {\tree};
\addplot[black, fill=cyan!99!white, mark=diamond*, mark size=5pt] table[x=cover,y=cart_acc_avg,col sep=comma] {\tree};

\end{axis}
\node[below=1.1cm of axis.south] {\leg};

\end{tikzpicture}%

    \hspace{-4pt}
  \def\file{./figs/adult_rf_avg.csv}%
  \def\tree{./figs/adult_tree.csv}%
  \def\ymax{0.87}%
  \def\ymin{0.80}%
  \def\leg{(c) adult}%
  \begin{tikzpicture}
\begin{axis}[
    name=axis,
    width=0.24\textwidth,
    xmin=0, xmax=1,
    ymin=\ymin, ymax=\ymax,
    xlabel={\footnotesize Transparency $T_m$},
    ylabel={\footnotesize Accuracy $A_m$},
    tick label style={font=\scriptsize},
    x label style={at={(axis description cs:0.5,0.05)},anchor=north},
    y label style={at={(axis description cs:0.2,0.5)},anchor=south},
]
\addplot[name path=companion_upper,draw=none,forget plot] table[x=cover,y expr=\thisrow{companion_acc_avg}+\thisrow{companion_acc_std}, col sep=comma] {\file};
\addplot[name path=companion_lower,draw=none,forget plot] table[x=cover,y expr=\thisrow{companion_acc_avg}-\thisrow{companion_acc_std}, col sep=comma] {\file};
\addplot [fill=blue!30,opacity=0.5,forget plot] fill between[of=companion_upper and companion_lower];

\addplot[name path=corels_upper,draw=none,forget plot] table[x=cover,y expr=\thisrow{corels_acc_avg}+\thisrow{corels_acc_std}, col sep=comma] {\file};
\addplot[name path=corels_lower,draw=none,forget plot] table[x=cover,y expr=\thisrow{corels_acc_avg}-\thisrow{corels_acc_std}, col sep=comma] {\file};
\addplot [fill=red!30,opacity=0.5,forget plot] fill between[of=corels_upper and corels_lower];

\addplot[name path=sbrl_upper,draw=none,forget plot] table[x=cover,y expr=\thisrow{sbrl_acc_avg}+\thisrow{sbrl_acc_std}, col sep=comma] {\file};
\addplot[name path=sbrl_lower,draw=none,forget plot] table[x=cover,y expr=\thisrow{sbrl_acc_avg}-\thisrow{sbrl_acc_std}, col sep=comma] {\file};
\addplot [fill=green!30,opacity=0.5,forget plot] fill between[of=sbrl_upper and sbrl_lower];

\addplot[blue, solid, line width=2pt] table[x=cover,y=companion_acc_avg,col sep=comma] {\file};
\addplot[red, dashed, line width=2pt] table[x=cover,y=corels_acc_avg,col sep=comma] {\file};
\addplot[green!40!black!60!, densely dotted, line width=2pt] table[x=cover,y=sbrl_acc_avg,col sep=comma] {\file};

\addplot[black, fill=magenta!99!white, mark=square*, mark size=4pt] table[x=cover,y=c45_acc_avg,col sep=comma] {\tree};
\addplot[black, fill=cyan!99!white, mark=diamond*, mark size=5pt] table[x=cover,y=cart_acc_avg,col sep=comma] {\tree};

\end{axis}
\node[below=1.1cm of axis.south] {\leg};

\end{tikzpicture}%

    \hspace{-4pt}
  \def\file{./figs/juvenile_rf_avg.csv}%
  \def\tree{./figs/juvenile_tree.csv}%
  \def\ymax{0.91}%
  \def\ymin{0.86}%
  \def\leg{(d) juvenile}%
  \begin{tikzpicture}
\begin{axis}[
    name=axis,
    width=0.24\textwidth,
    xmin=0, xmax=1,
    ymin=\ymin, ymax=\ymax,
    xlabel={\footnotesize Transparency $T_m$},
    ylabel={\footnotesize Accuracy $A_m$},
    tick label style={font=\scriptsize},
    x label style={at={(axis description cs:0.5,0.05)},anchor=north},
    y label style={at={(axis description cs:0.2,0.5)},anchor=south},
]
\addplot[name path=companion_upper,draw=none,forget plot] table[x=cover,y expr=\thisrow{companion_acc_avg}+\thisrow{companion_acc_std}, col sep=comma] {\file};
\addplot[name path=companion_lower,draw=none,forget plot] table[x=cover,y expr=\thisrow{companion_acc_avg}-\thisrow{companion_acc_std}, col sep=comma] {\file};
\addplot [fill=blue!30,opacity=0.5,forget plot] fill between[of=companion_upper and companion_lower];

\addplot[name path=corels_upper,draw=none,forget plot] table[x=cover,y expr=\thisrow{corels_acc_avg}+\thisrow{corels_acc_std}, col sep=comma] {\file};
\addplot[name path=corels_lower,draw=none,forget plot] table[x=cover,y expr=\thisrow{corels_acc_avg}-\thisrow{corels_acc_std}, col sep=comma] {\file};
\addplot [fill=red!30,opacity=0.5,forget plot] fill between[of=corels_upper and corels_lower];

\addplot[name path=sbrl_upper,draw=none,forget plot] table[x=cover,y expr=\thisrow{sbrl_acc_avg}+\thisrow{sbrl_acc_std}, col sep=comma] {\file};
\addplot[name path=sbrl_lower,draw=none,forget plot] table[x=cover,y expr=\thisrow{sbrl_acc_avg}-\thisrow{sbrl_acc_std}, col sep=comma] {\file};
\addplot [fill=green!30,opacity=0.5,forget plot] fill between[of=sbrl_upper and sbrl_lower];

\addplot[blue, solid, line width=2pt] table[x=cover,y=companion_acc_avg,col sep=comma] {\file};
\addplot[red, dashed, line width=2pt] table[x=cover,y=corels_acc_avg,col sep=comma] {\file};
\addplot[green!40!black!60!, densely dotted, line width=2pt] table[x=cover,y=sbrl_acc_avg,col sep=comma] {\file};

\addplot[black, fill=magenta!99!white, mark=square*, mark size=4pt] table[x=cover,y=c45_acc_avg,col sep=comma] {\tree};
\addplot[black, fill=cyan!99!white, mark=diamond*, mark size=5pt] table[x=cover,y=cart_acc_avg,col sep=comma] {\tree};

\end{axis}
\node[below=1.1cm of axis.south] {\leg};

\end{tikzpicture}%
\\
  \def\file{./figs/frisk_rf_avg.csv}%
  \def\tree{./figs/frisk_tree.csv}%
  \def\ymax{0.70}%
  \def\ymin{0.67}%
  \def\leg{(e) frisk}%
  \begin{tikzpicture}
\begin{axis}[
    name=axis,
    width=0.24\textwidth,
    xmin=0, xmax=1,
    ymin=\ymin, ymax=\ymax,
    xlabel={\footnotesize Transparency $T_m$},
    ylabel={\footnotesize Accuracy $A_m$},
    tick label style={font=\scriptsize},
    x label style={at={(axis description cs:0.5,0.05)},anchor=north},
    y label style={at={(axis description cs:0.2,0.5)},anchor=south},
]
\addplot[name path=companion_upper,draw=none,forget plot] table[x=cover,y expr=\thisrow{companion_acc_avg}+\thisrow{companion_acc_std}, col sep=comma] {\file};
\addplot[name path=companion_lower,draw=none,forget plot] table[x=cover,y expr=\thisrow{companion_acc_avg}-\thisrow{companion_acc_std}, col sep=comma] {\file};
\addplot [fill=blue!30,opacity=0.5,forget plot] fill between[of=companion_upper and companion_lower];

\addplot[name path=corels_upper,draw=none,forget plot] table[x=cover,y expr=\thisrow{corels_acc_avg}+\thisrow{corels_acc_std}, col sep=comma] {\file};
\addplot[name path=corels_lower,draw=none,forget plot] table[x=cover,y expr=\thisrow{corels_acc_avg}-\thisrow{corels_acc_std}, col sep=comma] {\file};
\addplot [fill=red!30,opacity=0.5,forget plot] fill between[of=corels_upper and corels_lower];

\addplot[name path=sbrl_upper,draw=none,forget plot] table[x=cover,y expr=\thisrow{sbrl_acc_avg}+\thisrow{sbrl_acc_std}, col sep=comma] {\file};
\addplot[name path=sbrl_lower,draw=none,forget plot] table[x=cover,y expr=\thisrow{sbrl_acc_avg}-\thisrow{sbrl_acc_std}, col sep=comma] {\file};
\addplot [fill=green!30,opacity=0.5,forget plot] fill between[of=sbrl_upper and sbrl_lower];

\addplot[blue, solid, line width=2pt] table[x=cover,y=companion_acc_avg,col sep=comma] {\file};
\addplot[red, dashed, line width=2pt] table[x=cover,y=corels_acc_avg,col sep=comma] {\file};
\addplot[green!40!black!60!, densely dotted, line width=2pt] table[x=cover,y=sbrl_acc_avg,col sep=comma] {\file};

\addplot[black, fill=magenta!99!white, mark=square*, mark size=4pt] table[x=cover,y=c45_acc_avg,col sep=comma] {\tree};
\addplot[black, fill=cyan!99!white, mark=diamond*, mark size=5pt] table[x=cover,y=cart_acc_avg,col sep=comma] {\tree};

\end{axis}
\node[below=1.1cm of axis.south] {\leg};

\end{tikzpicture}%

    \hspace{-4pt}
  \def\file{./figs/recidivism_rf_avg.csv}%
  \def\tree{./figs/recidivism_tree.csv}%
  \def\ymax{0.78}%
  \def\ymin{0.70}%
  \def\leg{(f) recidivism}%
  \begin{tikzpicture}
\begin{axis}[
    name=axis,
    width=0.24\textwidth,
    xmin=0, xmax=1,
    ymin=\ymin, ymax=\ymax,
    xlabel={\footnotesize Transparency $T_m$},
    ylabel={\footnotesize Accuracy $A_m$},
    tick label style={font=\scriptsize},
    x label style={at={(axis description cs:0.5,0.05)},anchor=north},
    y label style={at={(axis description cs:0.2,0.5)},anchor=south},
]
\addplot[name path=companion_upper,draw=none,forget plot] table[x=cover,y expr=\thisrow{companion_acc_avg}+\thisrow{companion_acc_std}, col sep=comma] {\file};
\addplot[name path=companion_lower,draw=none,forget plot] table[x=cover,y expr=\thisrow{companion_acc_avg}-\thisrow{companion_acc_std}, col sep=comma] {\file};
\addplot [fill=blue!30,opacity=0.5,forget plot] fill between[of=companion_upper and companion_lower];

\addplot[name path=corels_upper,draw=none,forget plot] table[x=cover,y expr=\thisrow{corels_acc_avg}+\thisrow{corels_acc_std}, col sep=comma] {\file};
\addplot[name path=corels_lower,draw=none,forget plot] table[x=cover,y expr=\thisrow{corels_acc_avg}-\thisrow{corels_acc_std}, col sep=comma] {\file};
\addplot [fill=red!30,opacity=0.5,forget plot] fill between[of=corels_upper and corels_lower];

\addplot[name path=sbrl_upper,draw=none,forget plot] table[x=cover,y expr=\thisrow{sbrl_acc_avg}+\thisrow{sbrl_acc_std}, col sep=comma] {\file};
\addplot[name path=sbrl_lower,draw=none,forget plot] table[x=cover,y expr=\thisrow{sbrl_acc_avg}-\thisrow{sbrl_acc_std}, col sep=comma] {\file};
\addplot [fill=green!30,opacity=0.5,forget plot] fill between[of=sbrl_upper and sbrl_lower];

\addplot[blue, solid, line width=2pt] table[x=cover,y=companion_acc_avg,col sep=comma] {\file};
\addplot[red, dashed, line width=2pt] table[x=cover,y=corels_acc_avg,col sep=comma] {\file};
\addplot[green!40!black!60!, densely dotted, line width=2pt] table[x=cover,y=sbrl_acc_avg,col sep=comma] {\file};

\addplot[black, fill=magenta!99!white, mark=square*, mark size=4pt] table[x=cover,y=c45_acc_avg,col sep=comma] {\tree};
\addplot[black, fill=cyan!99!white, mark=diamond*, mark size=5pt] table[x=cover,y=cart_acc_avg,col sep=comma] {\tree};

\end{axis}
\node[below=1.1cm of axis.south] {\leg};

\end{tikzpicture}%

    \hspace{-4pt}
  \def\file{./figs/magic_rf_avg.csv}%
  \def\tree{./figs/magic_tree.csv}%
  \def\ymax{0.89}%
  \def\ymin{0.80}%
  \def\leg{(g) magic}%
  \begin{tikzpicture}
\begin{axis}[
    name=axis,
    width=0.24\textwidth,
    xmin=0, xmax=1,
    ymin=\ymin, ymax=\ymax,
    xlabel={\footnotesize Transparency $T_m$},
    ylabel={\footnotesize Accuracy $A_m$},
    tick label style={font=\scriptsize},
    x label style={at={(axis description cs:0.5,0.05)},anchor=north},
    y label style={at={(axis description cs:0.2,0.5)},anchor=south},
]
\addplot[name path=companion_upper,draw=none,forget plot] table[x=cover,y expr=\thisrow{companion_acc_avg}+\thisrow{companion_acc_std}, col sep=comma] {\file};
\addplot[name path=companion_lower,draw=none,forget plot] table[x=cover,y expr=\thisrow{companion_acc_avg}-\thisrow{companion_acc_std}, col sep=comma] {\file};
\addplot [fill=blue!30,opacity=0.5,forget plot] fill between[of=companion_upper and companion_lower];

\addplot[name path=corels_upper,draw=none,forget plot] table[x=cover,y expr=\thisrow{corels_acc_avg}+\thisrow{corels_acc_std}, col sep=comma] {\file};
\addplot[name path=corels_lower,draw=none,forget plot] table[x=cover,y expr=\thisrow{corels_acc_avg}-\thisrow{corels_acc_std}, col sep=comma] {\file};
\addplot [fill=red!30,opacity=0.5,forget plot] fill between[of=corels_upper and corels_lower];

\addplot[name path=sbrl_upper,draw=none,forget plot] table[x=cover,y expr=\thisrow{sbrl_acc_avg}+\thisrow{sbrl_acc_std}, col sep=comma] {\file};
\addplot[name path=sbrl_lower,draw=none,forget plot] table[x=cover,y expr=\thisrow{sbrl_acc_avg}-\thisrow{sbrl_acc_std}, col sep=comma] {\file};
\addplot [fill=green!30,opacity=0.5,forget plot] fill between[of=sbrl_upper and sbrl_lower];

\addplot[blue, solid, line width=2pt] table[x=cover,y=companion_acc_avg,col sep=comma] {\file};
\addplot[red, dashed, line width=2pt] table[x=cover,y=corels_acc_avg,col sep=comma] {\file};
\addplot[green!40!black!60!, densely dotted, line width=2pt] table[x=cover,y=sbrl_acc_avg,col sep=comma] {\file};

\addplot[black, fill=magenta!99!white, mark=square*, mark size=4pt] table[x=cover,y=c45_acc_avg,col sep=comma] {\tree};
\addplot[black, fill=cyan!99!white, mark=diamond*, mark size=5pt] table[x=cover,y=cart_acc_avg,col sep=comma] {\tree};

\end{axis}
\node[below=1.1cm of axis.south] {\leg};

\end{tikzpicture}%

    \hspace{-4pt}
  \def\file{./figs/coupon_rf_avg.csv}%
  \def\tree{./figs/coupon_tree.csv}%
  \def\ymax{0.80}%
  \def\ymin{0.65}%
  \def\leg{(h) coupon}%
  \begin{tikzpicture}
\begin{axis}[
    name=axis,
    width=0.24\textwidth,
    xmin=0, xmax=1,
    ymin=\ymin, ymax=\ymax,
    xlabel={\footnotesize Transparency $T_m$},
    ylabel={\footnotesize Accuracy $A_m$},
    tick label style={font=\scriptsize},
    x label style={at={(axis description cs:0.5,0.05)},anchor=north},
    y label style={at={(axis description cs:0.2,0.5)},anchor=south},
]
\addplot[name path=companion_upper,draw=none,forget plot] table[x=cover,y expr=\thisrow{companion_acc_avg}+\thisrow{companion_acc_std}, col sep=comma] {\file};
\addplot[name path=companion_lower,draw=none,forget plot] table[x=cover,y expr=\thisrow{companion_acc_avg}-\thisrow{companion_acc_std}, col sep=comma] {\file};
\addplot [fill=blue!30,opacity=0.5,forget plot] fill between[of=companion_upper and companion_lower];

\addplot[name path=corels_upper,draw=none,forget plot] table[x=cover,y expr=\thisrow{corels_acc_avg}+\thisrow{corels_acc_std}, col sep=comma] {\file};
\addplot[name path=corels_lower,draw=none,forget plot] table[x=cover,y expr=\thisrow{corels_acc_avg}-\thisrow{corels_acc_std}, col sep=comma] {\file};
\addplot [fill=red!30,opacity=0.5,forget plot] fill between[of=corels_upper and corels_lower];

\addplot[name path=sbrl_upper,draw=none,forget plot] table[x=cover,y expr=\thisrow{sbrl_acc_avg}+\thisrow{sbrl_acc_std}, col sep=comma] {\file};
\addplot[name path=sbrl_lower,draw=none,forget plot] table[x=cover,y expr=\thisrow{sbrl_acc_avg}-\thisrow{sbrl_acc_std}, col sep=comma] {\file};
\addplot [fill=green!30,opacity=0.5,forget plot] fill between[of=sbrl_upper and sbrl_lower];

\addplot[blue, solid, line width=2pt] table[x=cover,y=companion_acc_avg,col sep=comma] {\file};
\addplot[red, dashed, line width=2pt] table[x=cover,y=corels_acc_avg,col sep=comma] {\file};
\addplot[green!40!black!60!, densely dotted, line width=2pt] table[x=cover,y=sbrl_acc_avg,col sep=comma] {\file};

\addplot[black, fill=magenta!99!white, mark=square*, mark size=4pt] table[x=cover,y=c45_acc_avg,col sep=comma] {\tree};
\addplot[black, fill=cyan!99!white, mark=diamond*, mark size=5pt] table[x=cover,y=cart_acc_avg,col sep=comma] {\tree};

\end{axis}
\node[below=1.1cm of axis.south] {\leg};

\end{tikzpicture}%

    \caption{The transparency--accuracy trade-off (with RandomForest as $f_b$): The solid lines denote average trade-off curves, while the shaded regions denote $\pm$ standard deviations evaluated via 5-fold cross validation.}
    \label{fig:rf}
\end{figure*}

\vspace{-2mm}
\paragraph{Setup}
To obtain a black-box $f_b$, we train RandomForest~\cite{breiman2001random}, AdaBoost~\cite{freund1997decision}, and XGBoost~\cite{chen2016xgboost}.
For the rule mining in Algorithm~\ref{alg:A}, we set the caridinality of each rule to be two and minimal support to be $0.05$ so that rules capturing too few observations are eliminated.
We also set the temperature $C_0 = 0.001$ and the training iteration $N = 50,000$.
For all the models, we tuned their hyper-parameters so that the maximum number of rules to be less than 20.\footnote{See Appendix~\ref{app:param} for the detailed parameter setups.}
All the models were trained and tested on a 5-fold cross validation.
In each fold, we trained all the models using 80\% of the data as train set, and evaluated their transparencies and the accuracies on the held out 20\% test set.

\vspace{-2mm}
\paragraph{Result}
We show the average transparency--accuracy trade-off curves for the eight data sets in \figurename~\ref{fig:rf}.
Because the results are similar for all the black-box models, we selected RandomForest.\footnote{See Appendix~\ref{app:result} for AdaBoost and XGBoost.}
In the figures, for each of CRL, CORELS, and SBRL, we draw the average of the 5-fold as solid lines and the standard deviation as the shaded regions.
The horizontal axis represents the transparency, and the vertical axis represents the accuracy.
The average accuracies of CART and C4.5 are shown as markers.

It is clearly observed that the curves of CRL are the highest in the figures for almost all data sets.
%This implies that the proposed CRL could provide the users with a flexible choice of the output with high accuracies.
This implies that CRL could provide the users with a flexible choice of the output with high accuracies.
The lower curves of CORELS and SBRL indicate that these models are suboptimal because they did not collaborate with the black-box model $f_b$ during training.
It is also interesting to observe that CRL has comparable performance as other rule-based models such as CART and C4.5 when reaching transparency of 100\%.
That is, the use of CRL as a stand-alone rule list is still a solid choice for the users.
Thus, CRL alone satisfies users' need to go from transparency of 0\% to 100\%.

\begin{table}[h]
\vspace{-6pt}
\caption{AUTACs for all models. The numbers in the parenthesis denote standard deviations. Bold fonts denote the best results (underlined), and the results which was not significantly different from the best result (t-test with the 5\% significance level).}
\label{tab:auc_rf}
\small
\begin{tabular}{llll}
\toprule
{} &                             CRL &             CORELS    &             SBRL         \\
\midrule
messidor   &  \underline{\textbf{.688 (.016)}} &  \textbf{.682 (.022)} &  \textbf{.669 (.026)} \\
german     & \underline{\textbf{.749 (.013)}} &  \textbf{.736 (.012)} &  \textbf{.742 (.010)} \\
adult      &  \underline{\textbf{.851 (.001)}} &           .835 (.004) &  \textbf{.847 (.005)} \\
juvenile   &  \underline{\textbf{.898 (.008)}} &  \textbf{.887 (.014)} &           .884 (.005) \\
frisk      &  \underline{\textbf{.688 (.003)}} &           .684 (.003) &           .681 (.003) \\
recidivism &  \underline{\textbf{.761 (.004)}} &           .742 (.006) &           .757 (.005) \\
magic      &  \underline{\textbf{.865 (.006)}} &           .842 (.006) &           .854 (.007) \\
coupon     &  \textbf{.740 (.014)} &           .716 (.017) &  \underline{\textbf{.742 (.021)}} \\
\bottomrule
\end{tabular}
%\vspace{-12pt}
\end{table}\normalsize

\tablename~\ref{tab:auc_rf} summarizes AUTACs on all the data sets.
%The table shows the clear advantage of the proposed CRL that it successfully attained the better trade-off over CORELS and SBRL.
The table shows the clear advantage of CRL that it successfully attained the better trade-off over CORELS and SBRL.
This result confirms the effectiveness of the proposed training algorithm that can seek for a good collaboration of the rule list and the black-box.

\vspace{-4mm}
\paragraph{An example}

\begin{table*}[h]
\vspace{-12pt}
\caption{An example of CRL on juvenile dataset (BLX represents black-box model)}\label{tab:juvenile}
\small
\begin{tabular}{p{35pt}p{305pt}p{10pt}p{20pt}p{20pt}p{20pt}}
\toprule
&\multicolumn{1}{c}{Companion Rule}                       & $\hat{y}$ & Rule cover & Rule acc. & BLX acc. \\ \hline
\textbf{if} & ``Have your friends ever broken into a vehicle or building to steal something" = ``No" \textbf{and} ``Has anyone-including your family members or friends - ever attacked you with a gun, knife or some other weapon?"=``Yes"    & 0          & 83.7\%      & 93.3\%    & 93.3\%   \\
\textbf{else if} &  Sex = Male \textbf{and} ``Have you ever tried cigarette smoking, even one or two puffs?" = ``Yes"        & 1          & 6.6\%      & 66.0\%     & 69.7\%   \\
\textbf{else if} &  ``Have your friends ever used prescription drugs such as amphetamines or barbiturates when there was no medical need for them" = ``No" \textbf{and} hard drug = 0 & 0          & 6.8\%      & 61.8\%    & 67.3\%   \\
\textbf{else if} &  ``Have your friends ever used alcohol" \textbf{and} Witness violence = TRUE & 1    & 2.6\%     & 52.4\%    & 71.4\%\\ \bottomrule
\end{tabular}
\end{table*}\normalsize In Table \ref{tab:juvenile}, we show an example of CRL model trained on the juvenile dataset, which predicts whether a child will commit delinquency, based on his prior growing-up experience. The data was collected via a survey where a child was asked to provide information about his prior exposure to violence from his friends, family, or community.

%%%%%%%%%%%%%%%%%%%%%%%%%%%%%%%%%%%%%%%%%%%%%%%%%%%%%%%%%%%%%%%%%%%%%%%%
%\vspace{-2mm}
\subsection{Human Evaluation of Transparency--Accuracy Trade-off}

We evaluate the trade-off between transparency and accuracy for humans, specifically, how much is a person willing to sacrifice accuracy for transparency. 

For this purpose, we designed a survey where we showed CRL models learned from three datasets, adult, german, and recidivism.
For each prediction, we showed two options, a companion rule and a black-box model, and the estimated accuracy for both of them.
Then we asked the participants to choose one from them.
See the Appendix~\ref{app:survey} for an example of the questions we designed.
We designed 18 questions and randomly showed 12 of them for each participant.

We collected responses from \x ~subjects in total and removed ten responses that fail the validation questions.
The average age of the participants was 28, from the youngest 20 to the oldest 68.
41\% of them were male.
The majority of the participants were undergraduate and graduate students from the computer science department and business schools, who are current and future users of machine learning models.
The rest were researchers from different domains, including business, medicine, and pharmacy, where interpretability is highly appreciated.

\begin{figure}[h]
    \centering
    \begin{tikzpicture}
    \begin{axis}[
        width=0.4\textwidth,
        height=0.27\textwidth,
        xmin=-0.25, xmax=0.05,
        ymin=0.2, ymax=1.0,
        xlabel={\footnotesize Relative accuracy loss if using a rule},
        ylabel style={align=center},
        ylabel={\footnotesize Ratios of the participants\\ preferred a rule},
        xtick={-0.25,-0.2,-0.15,-0.10,-0.05,0.0,0.05},
        xticklabel style={
        	/pgf/number format/fixed,
            /pgf/number format/precision=2,
            /pgf/number format/fixed zerofill
        },
        tick label style={font=\scriptsize},
        scatter/classes={a={mark=o,draw=blue,fill=blue}},
    ]
    \addplot[only marks,mark=*,draw=blue,fill=blue]
    table[meta=label] {
    x y label
-0.008290155440414507 0.8571428571428571 a
-0.012282497441146366 0.8 a
-0.0735981308411215 0.36585365853658536 a
-0.041343669250646 0.7560975609756098 a
-0.029126213592233007 0.7804878048780488 a
-0.08502024291497975 0.5135135135135135 a
-0.2402496099843994 0.275 a
0.0339943342776204 1.0 a
0.006544502617801046 1.0 a
-0.0059031877213695395 0.851063829787234 a
-0.044733044733044736 0.5909090909090909 a
-0.14027777777777778 0.5227272727272727 a
-0.2730109204368175 0.2558139534883721 a
0.03443877551020408 1.0 a
-0.051696284329563816 0.6046511627906976 a
-0.026683608640406607 0.7317073170731707 a
0.0 0.9777777777777777 a
-0.06424242424242424 0.5454545454545454 a
    };
    \end{axis}
    \end{tikzpicture}
    \caption{Human evaluated trade-off between model transparency and accuracy}
    \label{fig:trade-off} 
\end{figure}
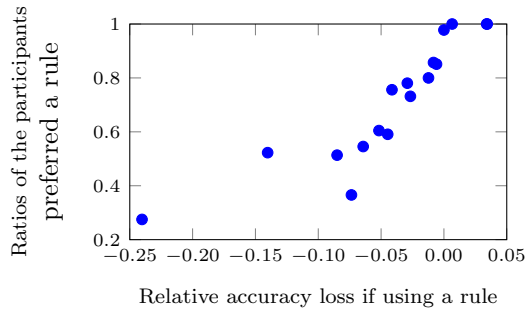

\figurename~\ref{fig:trade-off} reports the percentages of participants who chose a companion rule over the black-box at different loss of accuracy of using a rule.
The accuracy is reported in relative, i.e., the reduced accuracy for using a companion rule instead of the black-box divided by the accuracy of the black-box.
The relative reduction in accuracy ranged from -25\% to 5\%, with the positive being rules that were better than the black-box model and negative being the ones that were worse.
Results show that 80\% of the participants were tolerable up to 4\% accuracy loss or less.
Increasing that loss to 5\% lost about another 20\% of the participants.
\vspace{-3mm}

\paragraph{Findings} The results point out two interesting findings. First, while people desire interpretability, they are still very strict about accuracy. Interpretable models that lose too much accuracy are unlikely to be adopted in practice, which opens up opportunities for hybrid models like ours.
%Second, different users' preference for interpretability and accuracy can be so dramatically different: the curve covers a large range of accuracy drop while still having at least 30\% users who prefer an interpretable model even at the loss of 25\% of accuracy.\footnote{We also observed that the younger group are more likely to choose interpretable rules over black-box models than the older group. See Appendix~\ref{app:survey}.}.
%It is therefore important to leave the model selection at the end-user level, not the designer level. This is what CRL aims to do.
Second, different users' preference for interpretability and accuracy can be so dramatically different: the curve covers a large range of accuracy drop while still having at least 30\% users who prefer an interpretable model even at the loss of 25\% of accuracy. 
We also observed that the younger group are more likely to choose interpretable rules over black-box models than the older group.\footnote{See Appendix~\ref{app:survey} for the detail.}
It is therefore important to leave the model selection at the end-user level, not the designer level. This is what CRL aims to do.

%%%%%%%%%%%%%%%%%%%%%%%%%%%%%%%%%%%%%%%%%%%%%%%%%%%%%%%%%%%%%%%%%%%%%%%%
%\vspace{-2mm}
\section{Conclusion}
\label{sec:conclusion}

We proposed the companion rule list (CRL) for users to freely switch between the interpretable and the black-box models.
CRL is flexible enough so that users can choose whether to adopt the rule-based prediction or the black-box prediction for any of the input, based on their preferences on the interpretability and the accuracy.
We also designed a novel objective function, the area under the transparency--accuracy trade-off curve (AUTAC), for training a high-quality CRL, which we optimized by using the stochastic local search, utilizing information from the collaborative black-box model to form better collaboration between rules and the black-box model.
Our result confirms that this collaborative training yields better companion rules than those rule lists trained independently. 

Another main contribution of our paper is that we conducted an extensive human evaluation on \x~participants to study humans' views on model transparency versus accuracy. From the data we collected, we can understand how tolerable are humans to accuracy loss to gain model transparency. Results suggest that in practice, a stand-alone interpretable model will not be used by users if they drop too much in predictive performance. In addition, different people have a diverse preference for transparency, so the model selection should be made at the end-user level.  Both open up opportunities for companion models.

Lastly, we believe that combining an interpretable model with a black-box model is a promising framework that can easily trade-off between interpretability and accuracy.
%The current CRL will be suitable mostly for structured tabular data because the rules need to use semantic features.
%Extending the current framework of CRL to several data domains such as images, audios, and texts would be an important future direction.
While we mainly focused on tabular data in this study, extending the current framework of CRL to several data domains such as images, audios, and texts would be an important future direction.

%%%%%%%%%%%%%%%%%%%%%%%%%%%%%%%%%%%%%%%%%%%%%%%%%%%%%%%%%%%%%%%%%%%%%%%%
\subsubsection*{Acknowledgments}
Danqing Pan and Satoshi Hara are supported by JSPS KAKENHI Grant Number JP18K18471 and JP18K18106.

\bibliographystyle{unsrt}
\bibliography{companion}

\begin{thebibliography}{10}

\bibitem{wang2017bayesian}
Tong Wang, Cynthia Rudin, Finale Doshi-Velez, Yimin Liu, Erica Klampfl, and
  Perry MacNeille.
\newblock A bayesian framework for learning rule sets for interpretable
  classification.
\newblock {\em The Journal of Machine Learning Research}, 18(1):2357--2393,
  2017.

\bibitem{zeng2017interpretable}
Jiaming Zeng, Berk Ustun, and Cynthia Rudin.
\newblock Interpretable classification models for recidivism prediction.
\newblock {\em Journal of the Royal Statistical Society: Series A (Statistics
  in Society)}, 180(3):689--722, 2017.

\bibitem{ribeiro2016should}
Marco~Tulio Ribeiro, Sameer Singh, and Carlos Guestrin.
\newblock Why should {I} trust you?: Explaining the predictions of any
  classifier.
\newblock In {\em Proceedings of the 22nd ACM SIGKDD international conference
  on knowledge discovery and data mining}, pages 1135--1144. ACM, 2016.

\bibitem{lundberg2017unified}
Scott~M Lundberg and Su-In Lee.
\newblock A unified approach to interpreting model predictions.
\newblock In {\em Advances in Neural Information Processing Systems}, pages
  4765--4774, 2017.

\bibitem{rudin2019stop}
Cynthia Rudin.
\newblock Stop explaining black box machine learning models for high stakes
  decisions and use interpretable models instead.
\newblock {\em Nature Machine Intelligence}, 1(5):206, 2019.

\bibitem{wang2019hybrid}
Tong Wang and Qihang Lin.
\newblock Hybrid predictive model: When an interpretable model collaborates
  with a black-box model.
\newblock {\em arXiv preprint arXiv:1905.04241}, 2019.

\bibitem{wang2015trading}
Jialei Wang, Ryohei Fujimaki, and Yosuke Motohashi.
\newblock Trading interpretability for accuracy: Oblique treed sparse additive
  models.
\newblock In {\em Proceedings of the 21th ACM SIGKDD International Conference
  on Knowledge Discovery and Data Mining}, pages 1245--1254. ACM, 2015.

\bibitem{angelino2017learning}
Elaine Angelino, Nicholas Larus-Stone, Daniel Alabi, Margo Seltzer, and Cynthia
  Rudin.
\newblock Learning certifiably optimal rule lists.
\newblock In {\em Proceedings of the 23rd ACM SIGKDD International Conference
  on Knowledge Discovery and Data Mining}, pages 35--44. ACM, 2017.

\bibitem{yang2017scalable}
Hongyu Yang, Cynthia Rudin, and Margo Seltzer.
\newblock Scalable bayesian rule lists.
\newblock In {\em Proceedings of the 34th International Conference on Machine
  Learning}, pages 3921--3930, 2017.

\bibitem{ribeiro2018anchors}
Marco~Tulio Ribeiro, Sameer Singh, and Carlos Guestrin.
\newblock Anchors: High-precision model-agnostic explanations.
\newblock In {\em Thirty-Second AAAI Conference on Artificial Intelligence},
  2018.

\bibitem{guidotti2018local}
Riccardo Guidotti, Anna Monreale, Salvatore Ruggieri, Dino Pedreschi, Franco
  Turini, and Fosca Giannotti.
\newblock Local rule-based explanations of black box decision systems.
\newblock {\em arXiv preprint arXiv:1805.10820}, 2018.

\bibitem{aivodji2019fairwashing}
Ulrich A{\"\i}vodji, Hiromi Arai, Olivier Fortineau, S{\'e}bastien Gambs,
  Satoshi Hara, and Alain Tapp.
\newblock Fairwashing: the risk of rationalization.
\newblock In {\em Proceedings of the 36th International Conference on Machine
  Learning}, pages 161--170, 2019.

\bibitem{wang2018multi}
Tong Wang.
\newblock Multi-value rule sets for interpretable classification with
  feature-efficient representations.
\newblock In {\em Advances in Neural Information Processing Systems}, pages
  10835--10845, 2018.

\bibitem{dash2018boolean}
Sanjeeb Dash, Oktay Gunluk, and Dennis Wei.
\newblock Boolean decision rules via column generation.
\newblock In {\em Advances in Neural Information Processing Systems}, pages
  4655--4665, 2018.

\bibitem{lakkaraju2016interpretable}
Himabindu Lakkaraju, Stephen~H Bach, and Jure Leskovec.
\newblock Interpretable decision sets: A joint framework for description and
  prediction.
\newblock In {\em Proceedings of the 22nd ACM SIGKDD International Conference
  on Knowledge Discovery and Data Mining}, pages 1675--1684, 2016.

\bibitem{aivodji2019learning}
Ulrich A{\"\i}vodji, Julien Ferry, S{\'e}bastien Gambs, Marie-Jos{\'e} Huguet,
  and Mohamed Siala.
\newblock Learning fair rule lists.
\newblock {\em arXiv preprint arXiv:1909.03977}, 2019.

\bibitem{kirkpatrick1983optimization}
Scott Kirkpatrick, C~Daniel Gelatt, and Mario~P Vecchi.
\newblock Optimization by simulated annealing.
\newblock {\em science}, 220(4598):671--680, 1983.

\bibitem{nan2017adaptive}
Feng Nan and Venkatesh Saligrama.
\newblock Adaptive classification for prediction under a budget.
\newblock In {\em Advances in Neural Information Processing Systems}, pages
  4727--4737, 2017.

\bibitem{wang2019gaining}
Tong Wang.
\newblock Gaining free or low-cost interpretability with interpretable partial
  substitute.
\newblock In {\em Proceedings of the 36th International Conference on Machine
  Learning}, pages 6505--6514, 2019.

\bibitem{han2000mining}
Jiawei Han, Jian Pei, and Yiwen Yin.
\newblock Mining frequent patterns without candidate generation.
\newblock In {\em ACM SIGMOD Record}, volume~29, pages 1--12. ACM, 2000.

\bibitem{Dua:2019}
Dheeru Dua and Casey Graff.
\newblock {UCI} machine learning repository, 2017.

\bibitem{quinlan2014c4}
J~Ross Quinlan.
\newblock {\em C4. 5: programs for machine learning}.
\newblock Elsevier, 2014.

\bibitem{breiman2017classification}
Leo Breiman.
\newblock {\em Classification and regression trees}.
\newblock Routledge, 2017.

\bibitem{breiman2001random}
Leo Breiman.
\newblock Random forests.
\newblock {\em Machine learning}, 45(1):5--32, 2001.

\bibitem{freund1997decision}
Yoav Freund and Robert~E Schapire.
\newblock A decision-theoretic generalization of on-line learning and an
  application to boosting.
\newblock {\em Journal of computer and system sciences}, 55(1):119--139, 1997.

\bibitem{chen2016xgboost}
Tianqi Chen and Carlos Guestrin.
\newblock {XGB}oost: A scalable tree boosting system.
\newblock In {\em Proceedings of the 22nd acm sigkdd international conference
  on knowledge discovery and data mining}, pages 785--794. ACM, 2016.

\end{thebibliography}

%%%%%%%%%%%%%%%%%%%%%%%%%%%%%%%%%%%%%%%%%%%%%%%%%%%%%%%%%%%%%%%%%%%%%%%%
\clearpage
\appendix

%%%%%%%%%%%%%%%%%%%%%%%%%%%%%%%%%%%%%%%%%%%%%%%%%%%%%%%%%%%%%%%%%%%%%%%%
\section{Parameter Setups}
\label{app:param}

We set the parameters of the proposed training algorithm, CORELS, and SBRL, as follows.

\paragraph{Proposed Algorithm}
We tuned the penalty parameter $\alpha$ in the training objective function (\ref{eq:obj}) based on Algorithm~\ref{alg:CRL_tuning}.
For all the data sets, we mined rules containing maximally two conditions.
When the 16 GB memory on our machine was found to be insufficient for rule mining, we mined rules on a subset of the data set by reducing the number of observations.
We searched for the optimal $\alpha$ from the candidates ranging from $10^{-4}$ to $10^{-2}$.
We determined the optimal $\alpha$ as the one that maximized the AUTAC on the training set $D$, under the constraint that number of conditions is less than $20$.

\begin{algorithm}
    \caption{CRL Tuning Strategy}
    \label{alg:CRL_tuning}
    \begin{algorithmic}
        \STATE {\texttt{//Initial setting}}
        \STATE {$I_{\max} \leftarrow 20$ \hspace{6pt} $\triangleright$\texttt{max rules}}
        \STATE {$C_{\max} \leftarrow 2$ \hspace{6pt} $\triangleright$\texttt{max conditions}}
        \STATE {$A \leftarrow [.01, .005, .001, .0008, .0005, .0002, .0001]$ \\ \hspace{6pt} $\triangleright$\texttt{candidates of $\alpha$}}
        \STATE {$\alpha_\mathrm{opt} \leftarrow 0$, $\mathrm{AUTAC}_\mathrm{opt} \leftarrow 0$ }
        \STATE { FP-Growth minimal support $\leftarrow 0.05$ }
        \vspace{5pt}
        \STATE{\texttt{//Tuning rule mining}}
        \STATE { $b \leftarrow |D|$ \hspace{6pt} $\triangleright$\texttt{\# of observations for mining}}
        \WHILE { memory insufficient }
            \STATE{$b \leftarrow \lfloor 0.9b \rfloor$}
            \STATE { mine rules on $b$ observations with FP-Growth}
        \ENDWHILE
        \vspace{5pt}
        \STATE { \texttt{//Tuning $\alpha$} }
        \FOR{$\alpha \in A$}
            \STATE{$M, \mathrm{AUTAC} \leftarrow \mathrm{train}(D, \alpha, C_{\max})$}
            \IF{$M < I_{\max}$}
                \IF{$\mathrm{AUTAC} > \mathrm{AUTAC}_\mathrm{opt}$}
                    \STATE{$\mathrm{AUTAC}_\mathrm{opt} \leftarrow \mathrm{AUTAC}$}
                    \STATE{$\alpha_\mathrm{opt} \leftarrow \alpha$}
                \ENDIF
            \ENDIF

        \ENDFOR
        \STATE {\textbf{Output} $\alpha_\mathrm{opt}$}

\end{algorithmic}
\end{algorithm}

\newpage
\paragraph{CORELS}
For CORELS, we used an implementation publicly available at \url{https://github.com/fingoldin/pycorels}.
We tuned the maximal number of iterations $N$ and policy $P$ based on Algorithm~\ref{alg:CORELS_tuning}. We increment $N$ by $50k$ each time until $30k$ and search the best policy $P$ in list $L$. For all the data sets, we mined rules containing maximally two conditions. We determined the optimal $N$ and $P$ when predictive accuracy $\mathrm{ACC}$ is maximized on the testing set, under the constraint that condition number is less than 20.

\begin{algorithm}
    \caption{CORELS Tuning Strategy}
    \label{alg:CORELS_tuning}
    \begin{algorithmic}
        \STATE {\texttt{//Initial setting}}
        \STATE {$I_{\max} \leftarrow 20$ \hspace{6pt} $\triangleright$\texttt{max rules}}
        \STATE {$C_{\max} \leftarrow 2$ \hspace{6pt} $\triangleright$\texttt{max conditions}}
        \STATE {  $N \leftarrow $ 100k \hspace{6pt} $\triangleright$\texttt{maximum number of iterations}}
        \STATE { $ L \leftarrow \left [curious,lower bound,dfs,bfs,objective\right ] $ \\ \hspace{6pt} $\triangleright$\texttt{candidates of $P$}}
        \STATE { $N_\mathrm{opt} \leftarrow 0$, $\mathrm{ACC}_\mathrm{opt}  \leftarrow 0$, $P_\mathrm{opt} \leftarrow \mathrm{None}$}
        \vspace{5pt}
        \STATE{\texttt{//Tuning $N$ and $P$}}
        \FOR{$t = 1,2 ... 5$}
            \FOR{$P \in L$}
                \STATE{ $M, \mathrm{ACC} \leftarrow \mathrm{train}(D, N, P, C_{\max})$ }
                \IF{$M  <  I_{max}$}
                    \IF{$\mathrm{ACC}  > \mathrm{ACC}_\mathrm{opt} $}
                        \STATE{$\mathrm{ACC}_\mathrm{opt} \leftarrow \mathrm{ACC}$}
                        \STATE{$N_\mathrm{opt} \leftarrow N$}
                        \STATE{$P_\mathrm{opt} \leftarrow P$}
                    \ENDIF
                \ENDIF
            \ENDFOR
            \STATE { $N \leftarrow N + 50k$}
        \ENDFOR
        \STATE {\textbf{output} $N_\mathrm{opt}$, $P_\mathrm{opt}$}

\end{algorithmic}
\end{algorithm}

\newpage
\paragraph{SBRL}
For SBRL, we used an implementation publicly available at \url{https://github.com/Hongyuy/sbrlmod}.
We tuned number of chains $Nc$ and the expected length of the rule list $\lambda$ based on Algorithm~\ref{alg:SBRL_tuning}. We mine rules maximally containing two rules on messidor, german, adult, magic and coupon. When the 16 GB memory on our machine was found to be insufficient for rule mining, we increment both positive and negative minimal support $S_{+}$ and $S_{-}$ by $0.05$. When the minimal support is higher than an threshold $T$, we turn to mine rules maximally containing one rules. We mine rules maximally containing one rules on juvenile, frisk and recidivism. We determined the optimal $\lambda$ and $Nc$ as the ones that maximized predictive accuracy $\mathrm{ACC}$ on the testing set, under the constraint that number of conditions is less than 20.

\begin{algorithm}
    \caption{SBRL Tuning Strategy}
    \label{alg:SBRL_tuning}
    \begin{algorithmic}
        \STATE { \texttt{//Initial setting}}
        \STATE {$I_{\max} \leftarrow 20$ \hspace{6pt} $\triangleright$\texttt{max rules}}
        \STATE {$C_{\max} \leftarrow 2$ \hspace{6pt} $\triangleright$\texttt{max conditions}}
        \STATE { $T \leftarrow 0.7$ \hspace{6pt} $\triangleright$\texttt{threshold to reduce $C_{max}$} }
        \STATE { $\Lambda \leftarrow \left [ 1,2,5,10,15,20,25,30 \right ]$ \\ \hspace{6pt} $\triangleright$\texttt{candidates of $\lambda$}}
        \STATE { $Nc \leftarrow 0 \hspace{6pt} \triangleright \texttt{number of chains}$}
        \STATE { $\lambda_\mathrm{opt} \leftarrow 0$, $Nc_\mathrm{opt} \leftarrow 0$, $\mathrm{ACC}_\mathrm{opt} \leftarrow 0$ }
        \STATE{  $S_{+} \leftarrow 0.05$, $S_{-} \leftarrow 0.05 $ \\ 
        $\hspace{6pt} \triangleright$\texttt{minimal support for pos and neg rules} }
        \vspace{5pt}
        \STATE {\texttt{//Tuning minimal support}}
        \WHILE { memory insufficient}
            \STATE{ $S_{+} \leftarrow S_{+} + 0.05$ }
            \STATE{ $S_{-} \leftarrow S_{-} + 0.05$ }
            \IF { $S_{+} > T$ }
                \STATE{ $C_{\max} \leftarrow 1$ }
            \ENDIF
        \ENDWHILE
        \vspace{5pt}
        \STATE {\texttt{//Tuning $\lambda$ and $Nc$}}
        \FOR{ $\lambda \in \Lambda $ }
            \FOR{$t = 1 ... 6$}
                \STATE{$ M, \mathrm{ACC} \leftarrow \mathrm{train}(D, \lambda, Nc, S_{+}, S_{-}, C_{max})$ }
                \STATE { $Nc \leftarrow Nc + 5$ }
                \IF{ $M < I_{max}$ }
                    \IF{ $\mathrm{ACC} > \mathrm{ACC}_\mathrm{opt}$ }
                        \STATE{$\mathrm{ACC}_\mathrm{opt} \leftarrow \mathrm{ACC}$}
                        \STATE{ $\lambda_\mathrm{opt} \leftarrow \lambda$ }
                        \STATE{ $Nc_\mathrm{opt} \leftarrow Nc$ }
                    \ENDIF
                \ENDIF
            \ENDFOR
        \ENDFOR
               
        \STATE {\textbf{output} $\lambda_\mathrm{opt}$, $Nc_\mathrm{opt}$}

\end{algorithmic}
\end{algorithm}

%%%%%%%%%%%%%%%%%%%%%%%%%%%%%%%%%%%%%%%%%%%%%%%%%%%%%%%%%%%%%%%%%%%%%%%%
% \clearpage
\section{Exhaustive Results}
\label{app:result}

Here, we show the results for AdaBoost and XGBoost we omitted in Section~\ref{sec:experiment} due to space limitation.
Figures \ref{fig:ada} and \ref{fig:xg} show the trade-off curves, and Tables \ref{tab:auc_ada} and \ref{tab:auc_xg} show AUTACs.
These results also confirm the validity of the proposed training algorithm.

\begin{figure*}[h]
    \centering
    \begin{tikzpicture}
    \node[anchor=west] at (0,0) [draw,minimum width=11.2cm,minimum height=0.5cm] {};
    \node[anchor=west](crl) at (1.2, 0) {\footnotesize CRL};
    \draw[blue, solid, line width=2pt] (0.5, 0) -- (1.0, 0);
    \node[anchor=west, right=1.0 of crl](corels) {\footnotesize CORELS};
    \draw[red, dashed, line width=2pt] (2.5, 0) -- (3.0, 0);
    \node[anchor=west, right=1.0 of corels](sbrl) {\footnotesize SBRL};
    \draw[green!40!black!60!, densely dotted, line width=2pt] (5.1, 0) -- (5.6, 0);
    \node[anchor=west, right=1.0 of sbrl](c45) {\footnotesize C4.5};
    \draw[black,fill=magenta!99!white] (7.57,0.12) -- (7.57,-0.12) -- (7.33,-0.12) -- (7.33,0.12) -- cycle;
    \node[anchor=west, right=1.0 of c45](cart) {\footnotesize CART};
    \draw[black,fill=cyan!99!white] (9.35,0.15) -- (9.47,0.0) -- (9.35,-0.15) -- (9.23,0.0) -- cycle;
    \end{tikzpicture}\\
    \vspace{3pt}
  \def\file{./figs/messidor_ada_avg.csv}%
  \def\tree{./figs/messidor_tree.csv}%
  \def\ymax{0.75}%
  \def\ymin{0.60}%
  \def\leg{(a) messidor}%
  \begin{tikzpicture}
\begin{axis}[
    name=axis,
    width=0.24\textwidth,
    xmin=0, xmax=1,
    ymin=\ymin, ymax=\ymax,
    xlabel={\footnotesize Transparency $T_m$},
    ylabel={\footnotesize Accuracy $A_m$},
    tick label style={font=\scriptsize},
    x label style={at={(axis description cs:0.5,0.05)},anchor=north},
    y label style={at={(axis description cs:0.2,0.5)},anchor=south},
]
\addplot[name path=companion_upper,draw=none,forget plot] table[x=cover,y expr=\thisrow{companion_acc_avg}+\thisrow{companion_acc_std}, col sep=comma] {\file};
\addplot[name path=companion_lower,draw=none,forget plot] table[x=cover,y expr=\thisrow{companion_acc_avg}-\thisrow{companion_acc_std}, col sep=comma] {\file};
\addplot [fill=blue!30,opacity=0.5,forget plot] fill between[of=companion_upper and companion_lower];

\addplot[name path=corels_upper,draw=none,forget plot] table[x=cover,y expr=\thisrow{corels_acc_avg}+\thisrow{corels_acc_std}, col sep=comma] {\file};
\addplot[name path=corels_lower,draw=none,forget plot] table[x=cover,y expr=\thisrow{corels_acc_avg}-\thisrow{corels_acc_std}, col sep=comma] {\file};
\addplot [fill=red!30,opacity=0.5,forget plot] fill between[of=corels_upper and corels_lower];

\addplot[name path=sbrl_upper,draw=none,forget plot] table[x=cover,y expr=\thisrow{sbrl_acc_avg}+\thisrow{sbrl_acc_std}, col sep=comma] {\file};
\addplot[name path=sbrl_lower,draw=none,forget plot] table[x=cover,y expr=\thisrow{sbrl_acc_avg}-\thisrow{sbrl_acc_std}, col sep=comma] {\file};
\addplot [fill=green!30,opacity=0.5,forget plot] fill between[of=sbrl_upper and sbrl_lower];

\addplot[blue, solid, line width=2pt] table[x=cover,y=companion_acc_avg,col sep=comma] {\file};
\addplot[red, dashed, line width=2pt] table[x=cover,y=corels_acc_avg,col sep=comma] {\file};
\addplot[green!40!black!60!, densely dotted, line width=2pt] table[x=cover,y=sbrl_acc_avg,col sep=comma] {\file};

\addplot[black, fill=magenta!99!white, mark=square*, mark size=4pt] table[x=cover,y=c45_acc_avg,col sep=comma] {\tree};
\addplot[black, fill=cyan!99!white, mark=diamond*, mark size=5pt] table[x=cover,y=cart_acc_avg,col sep=comma] {\tree};

\end{axis}
\node[below=1.1cm of axis.south] {\leg};

\end{tikzpicture}%

    \hspace{-4pt}
  \def\file{./figs/german_ada_avg.csv}%
  \def\tree{./figs/german_tree.csv}%
  \def\ymax{0.78}%
  \def\ymin{0.68}%
  \def\leg{(b) german}%
  \begin{tikzpicture}
\begin{axis}[
    name=axis,
    width=0.24\textwidth,
    xmin=0, xmax=1,
    ymin=\ymin, ymax=\ymax,
    xlabel={\footnotesize Transparency $T_m$},
    ylabel={\footnotesize Accuracy $A_m$},
    tick label style={font=\scriptsize},
    x label style={at={(axis description cs:0.5,0.05)},anchor=north},
    y label style={at={(axis description cs:0.2,0.5)},anchor=south},
]
\addplot[name path=companion_upper,draw=none,forget plot] table[x=cover,y expr=\thisrow{companion_acc_avg}+\thisrow{companion_acc_std}, col sep=comma] {\file};
\addplot[name path=companion_lower,draw=none,forget plot] table[x=cover,y expr=\thisrow{companion_acc_avg}-\thisrow{companion_acc_std}, col sep=comma] {\file};
\addplot [fill=blue!30,opacity=0.5,forget plot] fill between[of=companion_upper and companion_lower];

\addplot[name path=corels_upper,draw=none,forget plot] table[x=cover,y expr=\thisrow{corels_acc_avg}+\thisrow{corels_acc_std}, col sep=comma] {\file};
\addplot[name path=corels_lower,draw=none,forget plot] table[x=cover,y expr=\thisrow{corels_acc_avg}-\thisrow{corels_acc_std}, col sep=comma] {\file};
\addplot [fill=red!30,opacity=0.5,forget plot] fill between[of=corels_upper and corels_lower];

\addplot[name path=sbrl_upper,draw=none,forget plot] table[x=cover,y expr=\thisrow{sbrl_acc_avg}+\thisrow{sbrl_acc_std}, col sep=comma] {\file};
\addplot[name path=sbrl_lower,draw=none,forget plot] table[x=cover,y expr=\thisrow{sbrl_acc_avg}-\thisrow{sbrl_acc_std}, col sep=comma] {\file};
\addplot [fill=green!30,opacity=0.5,forget plot] fill between[of=sbrl_upper and sbrl_lower];

\addplot[blue, solid, line width=2pt] table[x=cover,y=companion_acc_avg,col sep=comma] {\file};
\addplot[red, dashed, line width=2pt] table[x=cover,y=corels_acc_avg,col sep=comma] {\file};
\addplot[green!40!black!60!, densely dotted, line width=2pt] table[x=cover,y=sbrl_acc_avg,col sep=comma] {\file};

\addplot[black, fill=magenta!99!white, mark=square*, mark size=4pt] table[x=cover,y=c45_acc_avg,col sep=comma] {\tree};
\addplot[black, fill=cyan!99!white, mark=diamond*, mark size=5pt] table[x=cover,y=cart_acc_avg,col sep=comma] {\tree};

\end{axis}
\node[below=1.1cm of axis.south] {\leg};

\end{tikzpicture}%

    \hspace{-4pt}
  \def\file{./figs/adult_ada_avg.csv}%
  \def\tree{./figs/adult_tree.csv}%
  \def\ymax{0.87}%
  \def\ymin{0.80}%
  \def\leg{(c) adult}%
  \begin{tikzpicture}
\begin{axis}[
    name=axis,
    width=0.24\textwidth,
    xmin=0, xmax=1,
    ymin=\ymin, ymax=\ymax,
    xlabel={\footnotesize Transparency $T_m$},
    ylabel={\footnotesize Accuracy $A_m$},
    tick label style={font=\scriptsize},
    x label style={at={(axis description cs:0.5,0.05)},anchor=north},
    y label style={at={(axis description cs:0.2,0.5)},anchor=south},
]
\addplot[name path=companion_upper,draw=none,forget plot] table[x=cover,y expr=\thisrow{companion_acc_avg}+\thisrow{companion_acc_std}, col sep=comma] {\file};
\addplot[name path=companion_lower,draw=none,forget plot] table[x=cover,y expr=\thisrow{companion_acc_avg}-\thisrow{companion_acc_std}, col sep=comma] {\file};
\addplot [fill=blue!30,opacity=0.5,forget plot] fill between[of=companion_upper and companion_lower];

\addplot[name path=corels_upper,draw=none,forget plot] table[x=cover,y expr=\thisrow{corels_acc_avg}+\thisrow{corels_acc_std}, col sep=comma] {\file};
\addplot[name path=corels_lower,draw=none,forget plot] table[x=cover,y expr=\thisrow{corels_acc_avg}-\thisrow{corels_acc_std}, col sep=comma] {\file};
\addplot [fill=red!30,opacity=0.5,forget plot] fill between[of=corels_upper and corels_lower];

\addplot[name path=sbrl_upper,draw=none,forget plot] table[x=cover,y expr=\thisrow{sbrl_acc_avg}+\thisrow{sbrl_acc_std}, col sep=comma] {\file};
\addplot[name path=sbrl_lower,draw=none,forget plot] table[x=cover,y expr=\thisrow{sbrl_acc_avg}-\thisrow{sbrl_acc_std}, col sep=comma] {\file};
\addplot [fill=green!30,opacity=0.5,forget plot] fill between[of=sbrl_upper and sbrl_lower];

\addplot[blue, solid, line width=2pt] table[x=cover,y=companion_acc_avg,col sep=comma] {\file};
\addplot[red, dashed, line width=2pt] table[x=cover,y=corels_acc_avg,col sep=comma] {\file};
\addplot[green!40!black!60!, densely dotted, line width=2pt] table[x=cover,y=sbrl_acc_avg,col sep=comma] {\file};

\addplot[black, fill=magenta!99!white, mark=square*, mark size=4pt] table[x=cover,y=c45_acc_avg,col sep=comma] {\tree};
\addplot[black, fill=cyan!99!white, mark=diamond*, mark size=5pt] table[x=cover,y=cart_acc_avg,col sep=comma] {\tree};

\end{axis}
\node[below=1.1cm of axis.south] {\leg};

\end{tikzpicture}%

    \hspace{-4pt}
  \def\file{./figs/juvenile_ada_avg.csv}%
  \def\tree{./figs/juvenile_tree.csv}%
  \def\ymax{0.91}%
  \def\ymin{0.86}%
  \def\leg{(d) juvenile}%
  \begin{tikzpicture}
\begin{axis}[
    name=axis,
    width=0.24\textwidth,
    xmin=0, xmax=1,
    ymin=\ymin, ymax=\ymax,
    xlabel={\footnotesize Transparency $T_m$},
    ylabel={\footnotesize Accuracy $A_m$},
    tick label style={font=\scriptsize},
    x label style={at={(axis description cs:0.5,0.05)},anchor=north},
    y label style={at={(axis description cs:0.2,0.5)},anchor=south},
]
\addplot[name path=companion_upper,draw=none,forget plot] table[x=cover,y expr=\thisrow{companion_acc_avg}+\thisrow{companion_acc_std}, col sep=comma] {\file};
\addplot[name path=companion_lower,draw=none,forget plot] table[x=cover,y expr=\thisrow{companion_acc_avg}-\thisrow{companion_acc_std}, col sep=comma] {\file};
\addplot [fill=blue!30,opacity=0.5,forget plot] fill between[of=companion_upper and companion_lower];

\addplot[name path=corels_upper,draw=none,forget plot] table[x=cover,y expr=\thisrow{corels_acc_avg}+\thisrow{corels_acc_std}, col sep=comma] {\file};
\addplot[name path=corels_lower,draw=none,forget plot] table[x=cover,y expr=\thisrow{corels_acc_avg}-\thisrow{corels_acc_std}, col sep=comma] {\file};
\addplot [fill=red!30,opacity=0.5,forget plot] fill between[of=corels_upper and corels_lower];

\addplot[name path=sbrl_upper,draw=none,forget plot] table[x=cover,y expr=\thisrow{sbrl_acc_avg}+\thisrow{sbrl_acc_std}, col sep=comma] {\file};
\addplot[name path=sbrl_lower,draw=none,forget plot] table[x=cover,y expr=\thisrow{sbrl_acc_avg}-\thisrow{sbrl_acc_std}, col sep=comma] {\file};
\addplot [fill=green!30,opacity=0.5,forget plot] fill between[of=sbrl_upper and sbrl_lower];

\addplot[blue, solid, line width=2pt] table[x=cover,y=companion_acc_avg,col sep=comma] {\file};
\addplot[red, dashed, line width=2pt] table[x=cover,y=corels_acc_avg,col sep=comma] {\file};
\addplot[green!40!black!60!, densely dotted, line width=2pt] table[x=cover,y=sbrl_acc_avg,col sep=comma] {\file};

\addplot[black, fill=magenta!99!white, mark=square*, mark size=4pt] table[x=cover,y=c45_acc_avg,col sep=comma] {\tree};
\addplot[black, fill=cyan!99!white, mark=diamond*, mark size=5pt] table[x=cover,y=cart_acc_avg,col sep=comma] {\tree};

\end{axis}
\node[below=1.1cm of axis.south] {\leg};

\end{tikzpicture}%
\\
  \def\file{./figs/frisk_ada_avg.csv}%
  \def\tree{./figs/frisk_tree.csv}%
  \def\ymax{0.70}%
  \def\ymin{0.67}%
  \def\leg{(e) frisk}%
  \begin{tikzpicture}
\begin{axis}[
    name=axis,
    width=0.24\textwidth,
    xmin=0, xmax=1,
    ymin=\ymin, ymax=\ymax,
    xlabel={\footnotesize Transparency $T_m$},
    ylabel={\footnotesize Accuracy $A_m$},
    tick label style={font=\scriptsize},
    x label style={at={(axis description cs:0.5,0.05)},anchor=north},
    y label style={at={(axis description cs:0.2,0.5)},anchor=south},
]
\addplot[name path=companion_upper,draw=none,forget plot] table[x=cover,y expr=\thisrow{companion_acc_avg}+\thisrow{companion_acc_std}, col sep=comma] {\file};
\addplot[name path=companion_lower,draw=none,forget plot] table[x=cover,y expr=\thisrow{companion_acc_avg}-\thisrow{companion_acc_std}, col sep=comma] {\file};
\addplot [fill=blue!30,opacity=0.5,forget plot] fill between[of=companion_upper and companion_lower];

\addplot[name path=corels_upper,draw=none,forget plot] table[x=cover,y expr=\thisrow{corels_acc_avg}+\thisrow{corels_acc_std}, col sep=comma] {\file};
\addplot[name path=corels_lower,draw=none,forget plot] table[x=cover,y expr=\thisrow{corels_acc_avg}-\thisrow{corels_acc_std}, col sep=comma] {\file};
\addplot [fill=red!30,opacity=0.5,forget plot] fill between[of=corels_upper and corels_lower];

\addplot[name path=sbrl_upper,draw=none,forget plot] table[x=cover,y expr=\thisrow{sbrl_acc_avg}+\thisrow{sbrl_acc_std}, col sep=comma] {\file};
\addplot[name path=sbrl_lower,draw=none,forget plot] table[x=cover,y expr=\thisrow{sbrl_acc_avg}-\thisrow{sbrl_acc_std}, col sep=comma] {\file};
\addplot [fill=green!30,opacity=0.5,forget plot] fill between[of=sbrl_upper and sbrl_lower];

\addplot[blue, solid, line width=2pt] table[x=cover,y=companion_acc_avg,col sep=comma] {\file};
\addplot[red, dashed, line width=2pt] table[x=cover,y=corels_acc_avg,col sep=comma] {\file};
\addplot[green!40!black!60!, densely dotted, line width=2pt] table[x=cover,y=sbrl_acc_avg,col sep=comma] {\file};

\addplot[black, fill=magenta!99!white, mark=square*, mark size=4pt] table[x=cover,y=c45_acc_avg,col sep=comma] {\tree};
\addplot[black, fill=cyan!99!white, mark=diamond*, mark size=5pt] table[x=cover,y=cart_acc_avg,col sep=comma] {\tree};

\end{axis}
\node[below=1.1cm of axis.south] {\leg};

\end{tikzpicture}%

    \hspace{-4pt}
  \def\file{./figs/recidivism_ada_avg.csv}%
  \def\tree{./figs/recidivism_tree.csv}%
  \def\ymax{0.78}%
  \def\ymin{0.70}%
  \def\leg{(f) recidivism}%
  \begin{tikzpicture}
\begin{axis}[
    name=axis,
    width=0.24\textwidth,
    xmin=0, xmax=1,
    ymin=\ymin, ymax=\ymax,
    xlabel={\footnotesize Transparency $T_m$},
    ylabel={\footnotesize Accuracy $A_m$},
    tick label style={font=\scriptsize},
    x label style={at={(axis description cs:0.5,0.05)},anchor=north},
    y label style={at={(axis description cs:0.2,0.5)},anchor=south},
]
\addplot[name path=companion_upper,draw=none,forget plot] table[x=cover,y expr=\thisrow{companion_acc_avg}+\thisrow{companion_acc_std}, col sep=comma] {\file};
\addplot[name path=companion_lower,draw=none,forget plot] table[x=cover,y expr=\thisrow{companion_acc_avg}-\thisrow{companion_acc_std}, col sep=comma] {\file};
\addplot [fill=blue!30,opacity=0.5,forget plot] fill between[of=companion_upper and companion_lower];

\addplot[name path=corels_upper,draw=none,forget plot] table[x=cover,y expr=\thisrow{corels_acc_avg}+\thisrow{corels_acc_std}, col sep=comma] {\file};
\addplot[name path=corels_lower,draw=none,forget plot] table[x=cover,y expr=\thisrow{corels_acc_avg}-\thisrow{corels_acc_std}, col sep=comma] {\file};
\addplot [fill=red!30,opacity=0.5,forget plot] fill between[of=corels_upper and corels_lower];

\addplot[name path=sbrl_upper,draw=none,forget plot] table[x=cover,y expr=\thisrow{sbrl_acc_avg}+\thisrow{sbrl_acc_std}, col sep=comma] {\file};
\addplot[name path=sbrl_lower,draw=none,forget plot] table[x=cover,y expr=\thisrow{sbrl_acc_avg}-\thisrow{sbrl_acc_std}, col sep=comma] {\file};
\addplot [fill=green!30,opacity=0.5,forget plot] fill between[of=sbrl_upper and sbrl_lower];

\addplot[blue, solid, line width=2pt] table[x=cover,y=companion_acc_avg,col sep=comma] {\file};
\addplot[red, dashed, line width=2pt] table[x=cover,y=corels_acc_avg,col sep=comma] {\file};
\addplot[green!40!black!60!, densely dotted, line width=2pt] table[x=cover,y=sbrl_acc_avg,col sep=comma] {\file};

\addplot[black, fill=magenta!99!white, mark=square*, mark size=4pt] table[x=cover,y=c45_acc_avg,col sep=comma] {\tree};
\addplot[black, fill=cyan!99!white, mark=diamond*, mark size=5pt] table[x=cover,y=cart_acc_avg,col sep=comma] {\tree};

\end{axis}
\node[below=1.1cm of axis.south] {\leg};

\end{tikzpicture}%

    \hspace{-4pt}
  \def\file{./figs/magic_ada_avg.csv}%
  \def\tree{./figs/magic_tree.csv}%
  \def\ymax{0.89}%
  \def\ymin{0.80}%
  \def\leg{(g) magic}%
  \begin{tikzpicture}
\begin{axis}[
    name=axis,
    width=0.24\textwidth,
    xmin=0, xmax=1,
    ymin=\ymin, ymax=\ymax,
    xlabel={\footnotesize Transparency $T_m$},
    ylabel={\footnotesize Accuracy $A_m$},
    tick label style={font=\scriptsize},
    x label style={at={(axis description cs:0.5,0.05)},anchor=north},
    y label style={at={(axis description cs:0.2,0.5)},anchor=south},
]
\addplot[name path=companion_upper,draw=none,forget plot] table[x=cover,y expr=\thisrow{companion_acc_avg}+\thisrow{companion_acc_std}, col sep=comma] {\file};
\addplot[name path=companion_lower,draw=none,forget plot] table[x=cover,y expr=\thisrow{companion_acc_avg}-\thisrow{companion_acc_std}, col sep=comma] {\file};
\addplot [fill=blue!30,opacity=0.5,forget plot] fill between[of=companion_upper and companion_lower];

\addplot[name path=corels_upper,draw=none,forget plot] table[x=cover,y expr=\thisrow{corels_acc_avg}+\thisrow{corels_acc_std}, col sep=comma] {\file};
\addplot[name path=corels_lower,draw=none,forget plot] table[x=cover,y expr=\thisrow{corels_acc_avg}-\thisrow{corels_acc_std}, col sep=comma] {\file};
\addplot [fill=red!30,opacity=0.5,forget plot] fill between[of=corels_upper and corels_lower];

\addplot[name path=sbrl_upper,draw=none,forget plot] table[x=cover,y expr=\thisrow{sbrl_acc_avg}+\thisrow{sbrl_acc_std}, col sep=comma] {\file};
\addplot[name path=sbrl_lower,draw=none,forget plot] table[x=cover,y expr=\thisrow{sbrl_acc_avg}-\thisrow{sbrl_acc_std}, col sep=comma] {\file};
\addplot [fill=green!30,opacity=0.5,forget plot] fill between[of=sbrl_upper and sbrl_lower];

\addplot[blue, solid, line width=2pt] table[x=cover,y=companion_acc_avg,col sep=comma] {\file};
\addplot[red, dashed, line width=2pt] table[x=cover,y=corels_acc_avg,col sep=comma] {\file};
\addplot[green!40!black!60!, densely dotted, line width=2pt] table[x=cover,y=sbrl_acc_avg,col sep=comma] {\file};

\addplot[black, fill=magenta!99!white, mark=square*, mark size=4pt] table[x=cover,y=c45_acc_avg,col sep=comma] {\tree};
\addplot[black, fill=cyan!99!white, mark=diamond*, mark size=5pt] table[x=cover,y=cart_acc_avg,col sep=comma] {\tree};

\end{axis}
\node[below=1.1cm of axis.south] {\leg};

\end{tikzpicture}%

    \hspace{-4pt}
  \def\file{./figs/coupon_ada_avg.csv}%
  \def\tree{./figs/coupon_tree.csv}%
  \def\ymax{0.80}%
  \def\ymin{0.65}%
  \def\leg{(h) coupon}%
  \begin{tikzpicture}
\begin{axis}[
    name=axis,
    width=0.24\textwidth,
    xmin=0, xmax=1,
    ymin=\ymin, ymax=\ymax,
    xlabel={\footnotesize Transparency $T_m$},
    ylabel={\footnotesize Accuracy $A_m$},
    tick label style={font=\scriptsize},
    x label style={at={(axis description cs:0.5,0.05)},anchor=north},
    y label style={at={(axis description cs:0.2,0.5)},anchor=south},
]
\addplot[name path=companion_upper,draw=none,forget plot] table[x=cover,y expr=\thisrow{companion_acc_avg}+\thisrow{companion_acc_std}, col sep=comma] {\file};
\addplot[name path=companion_lower,draw=none,forget plot] table[x=cover,y expr=\thisrow{companion_acc_avg}-\thisrow{companion_acc_std}, col sep=comma] {\file};
\addplot [fill=blue!30,opacity=0.5,forget plot] fill between[of=companion_upper and companion_lower];

\addplot[name path=corels_upper,draw=none,forget plot] table[x=cover,y expr=\thisrow{corels_acc_avg}+\thisrow{corels_acc_std}, col sep=comma] {\file};
\addplot[name path=corels_lower,draw=none,forget plot] table[x=cover,y expr=\thisrow{corels_acc_avg}-\thisrow{corels_acc_std}, col sep=comma] {\file};
\addplot [fill=red!30,opacity=0.5,forget plot] fill between[of=corels_upper and corels_lower];

\addplot[name path=sbrl_upper,draw=none,forget plot] table[x=cover,y expr=\thisrow{sbrl_acc_avg}+\thisrow{sbrl_acc_std}, col sep=comma] {\file};
\addplot[name path=sbrl_lower,draw=none,forget plot] table[x=cover,y expr=\thisrow{sbrl_acc_avg}-\thisrow{sbrl_acc_std}, col sep=comma] {\file};
\addplot [fill=green!30,opacity=0.5,forget plot] fill between[of=sbrl_upper and sbrl_lower];

\addplot[blue, solid, line width=2pt] table[x=cover,y=companion_acc_avg,col sep=comma] {\file};
\addplot[red, dashed, line width=2pt] table[x=cover,y=corels_acc_avg,col sep=comma] {\file};
\addplot[green!40!black!60!, densely dotted, line width=2pt] table[x=cover,y=sbrl_acc_avg,col sep=comma] {\file};

\addplot[black, fill=magenta!99!white, mark=square*, mark size=4pt] table[x=cover,y=c45_acc_avg,col sep=comma] {\tree};
\addplot[black, fill=cyan!99!white, mark=diamond*, mark size=5pt] table[x=cover,y=cart_acc_avg,col sep=comma] {\tree};

\end{axis}
\node[below=1.1cm of axis.south] {\leg};

\end{tikzpicture}%

    \caption{The transparency--accuracy trade-off (with AdaBoost as $f_b$): The solid lines denote average trade-off curves, while the shaded regions denote $\pm$ standard deviations evaluated via 5-fold cross validation.}
    \label{fig:ada}
\end{figure*}
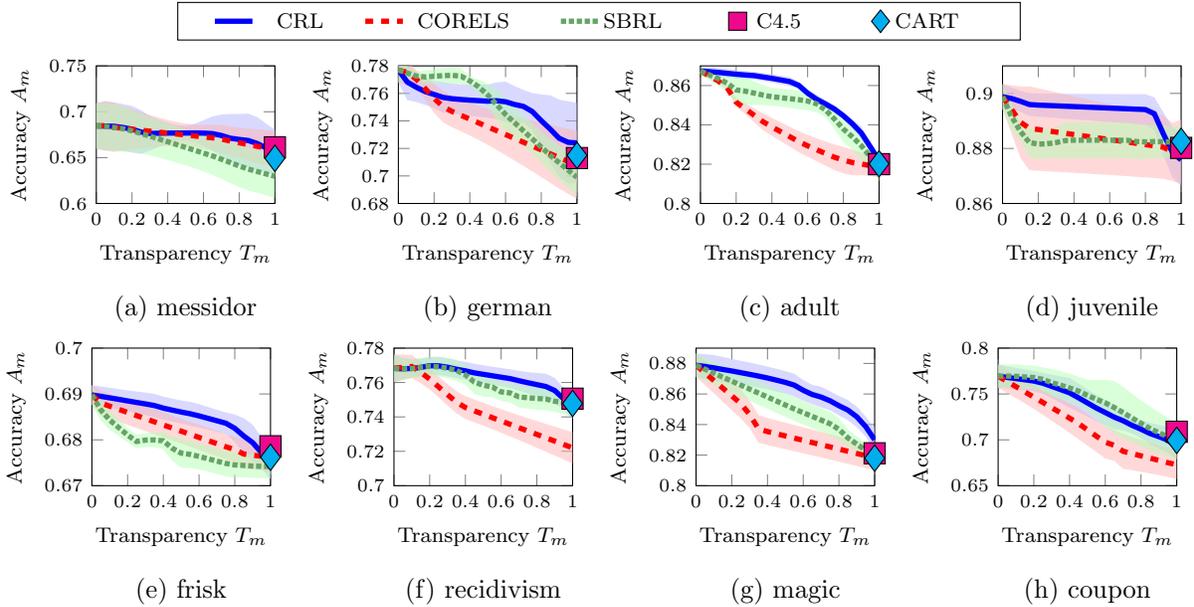

\begin{figure*}[h]
    \centering
    \begin{tikzpicture}
    \node[anchor=west] at (0,0) [draw,minimum width=11.2cm,minimum height=0.5cm] {};
    \node[anchor=west](crl) at (1.2, 0) {\footnotesize CRL};
    \draw[blue, solid, line width=2pt] (0.5, 0) -- (1.0, 0);
    \node[anchor=west, right=1.0 of crl](corels) {\footnotesize CORELS};
    \draw[red, dashed, line width=2pt] (2.5, 0) -- (3.0, 0);
    \node[anchor=west, right=1.0 of corels](sbrl) {\footnotesize SBRL};
    \draw[green!40!black!60!, densely dotted, line width=2pt] (5.1, 0) -- (5.6, 0);
    \node[anchor=west, right=1.0 of sbrl](c45) {\footnotesize C4.5};
    \draw[black,fill=magenta!99!white] (7.57,0.12) -- (7.57,-0.12) -- (7.33,-0.12) -- (7.33,0.12) -- cycle;
    \node[anchor=west, right=1.0 of c45](cart) {\footnotesize CART};
    \draw[black,fill=cyan!99!white] (9.35,0.15) -- (9.47,0.0) -- (9.35,-0.15) -- (9.23,0.0) -- cycle;
    \end{tikzpicture}\\
    \vspace{3pt}
  \def\file{./figs/messidor_xg_avg.csv}%
  \def\tree{./figs/messidor_tree.csv}%
  \def\ymax{0.75}%
  \def\ymin{0.60}%
  \def\leg{(a) messidor}%
  \begin{tikzpicture}
\begin{axis}[
    name=axis,
    width=0.24\textwidth,
    xmin=0, xmax=1,
    ymin=\ymin, ymax=\ymax,
    xlabel={\footnotesize Transparency $T_m$},
    ylabel={\footnotesize Accuracy $A_m$},
    tick label style={font=\scriptsize},
    x label style={at={(axis description cs:0.5,0.05)},anchor=north},
    y label style={at={(axis description cs:0.2,0.5)},anchor=south},
]
\addplot[name path=companion_upper,draw=none,forget plot] table[x=cover,y expr=\thisrow{companion_acc_avg}+\thisrow{companion_acc_std}, col sep=comma] {\file};
\addplot[name path=companion_lower,draw=none,forget plot] table[x=cover,y expr=\thisrow{companion_acc_avg}-\thisrow{companion_acc_std}, col sep=comma] {\file};
\addplot [fill=blue!30,opacity=0.5,forget plot] fill between[of=companion_upper and companion_lower];

\addplot[name path=corels_upper,draw=none,forget plot] table[x=cover,y expr=\thisrow{corels_acc_avg}+\thisrow{corels_acc_std}, col sep=comma] {\file};
\addplot[name path=corels_lower,draw=none,forget plot] table[x=cover,y expr=\thisrow{corels_acc_avg}-\thisrow{corels_acc_std}, col sep=comma] {\file};
\addplot [fill=red!30,opacity=0.5,forget plot] fill between[of=corels_upper and corels_lower];

\addplot[name path=sbrl_upper,draw=none,forget plot] table[x=cover,y expr=\thisrow{sbrl_acc_avg}+\thisrow{sbrl_acc_std}, col sep=comma] {\file};
\addplot[name path=sbrl_lower,draw=none,forget plot] table[x=cover,y expr=\thisrow{sbrl_acc_avg}-\thisrow{sbrl_acc_std}, col sep=comma] {\file};
\addplot [fill=green!30,opacity=0.5,forget plot] fill between[of=sbrl_upper and sbrl_lower];

\addplot[blue, solid, line width=2pt] table[x=cover,y=companion_acc_avg,col sep=comma] {\file};
\addplot[red, dashed, line width=2pt] table[x=cover,y=corels_acc_avg,col sep=comma] {\file};
\addplot[green!40!black!60!, densely dotted, line width=2pt] table[x=cover,y=sbrl_acc_avg,col sep=comma] {\file};

\addplot[black, fill=magenta!99!white, mark=square*, mark size=4pt] table[x=cover,y=c45_acc_avg,col sep=comma] {\tree};
\addplot[black, fill=cyan!99!white, mark=diamond*, mark size=5pt] table[x=cover,y=cart_acc_avg,col sep=comma] {\tree};

\end{axis}
\node[below=1.1cm of axis.south] {\leg};

\end{tikzpicture}%

    \hspace{-4pt}
  \def\file{./figs/german_xg_avg.csv}%
  \def\tree{./figs/german_tree.csv}%
  \def\ymax{0.78}%
  \def\ymin{0.68}%
  \def\leg{(b) german}%
  \begin{tikzpicture}
\begin{axis}[
    name=axis,
    width=0.24\textwidth,
    xmin=0, xmax=1,
    ymin=\ymin, ymax=\ymax,
    xlabel={\footnotesize Transparency $T_m$},
    ylabel={\footnotesize Accuracy $A_m$},
    tick label style={font=\scriptsize},
    x label style={at={(axis description cs:0.5,0.05)},anchor=north},
    y label style={at={(axis description cs:0.2,0.5)},anchor=south},
]
\addplot[name path=companion_upper,draw=none,forget plot] table[x=cover,y expr=\thisrow{companion_acc_avg}+\thisrow{companion_acc_std}, col sep=comma] {\file};
\addplot[name path=companion_lower,draw=none,forget plot] table[x=cover,y expr=\thisrow{companion_acc_avg}-\thisrow{companion_acc_std}, col sep=comma] {\file};
\addplot [fill=blue!30,opacity=0.5,forget plot] fill between[of=companion_upper and companion_lower];

\addplot[name path=corels_upper,draw=none,forget plot] table[x=cover,y expr=\thisrow{corels_acc_avg}+\thisrow{corels_acc_std}, col sep=comma] {\file};
\addplot[name path=corels_lower,draw=none,forget plot] table[x=cover,y expr=\thisrow{corels_acc_avg}-\thisrow{corels_acc_std}, col sep=comma] {\file};
\addplot [fill=red!30,opacity=0.5,forget plot] fill between[of=corels_upper and corels_lower];

\addplot[name path=sbrl_upper,draw=none,forget plot] table[x=cover,y expr=\thisrow{sbrl_acc_avg}+\thisrow{sbrl_acc_std}, col sep=comma] {\file};
\addplot[name path=sbrl_lower,draw=none,forget plot] table[x=cover,y expr=\thisrow{sbrl_acc_avg}-\thisrow{sbrl_acc_std}, col sep=comma] {\file};
\addplot [fill=green!30,opacity=0.5,forget plot] fill between[of=sbrl_upper and sbrl_lower];

\addplot[blue, solid, line width=2pt] table[x=cover,y=companion_acc_avg,col sep=comma] {\file};
\addplot[red, dashed, line width=2pt] table[x=cover,y=corels_acc_avg,col sep=comma] {\file};
\addplot[green!40!black!60!, densely dotted, line width=2pt] table[x=cover,y=sbrl_acc_avg,col sep=comma] {\file};

\addplot[black, fill=magenta!99!white, mark=square*, mark size=4pt] table[x=cover,y=c45_acc_avg,col sep=comma] {\tree};
\addplot[black, fill=cyan!99!white, mark=diamond*, mark size=5pt] table[x=cover,y=cart_acc_avg,col sep=comma] {\tree};

\end{axis}
\node[below=1.1cm of axis.south] {\leg};

\end{tikzpicture}%

    \hspace{-4pt}
  \def\file{./figs/adult_xg_avg.csv}%
  \def\tree{./figs/adult_tree.csv}%
  \def\ymax{0.87}%
  \def\ymin{0.80}%
  \def\leg{(c) adult}%
  \begin{tikzpicture}
\begin{axis}[
    name=axis,
    width=0.24\textwidth,
    xmin=0, xmax=1,
    ymin=\ymin, ymax=\ymax,
    xlabel={\footnotesize Transparency $T_m$},
    ylabel={\footnotesize Accuracy $A_m$},
    tick label style={font=\scriptsize},
    x label style={at={(axis description cs:0.5,0.05)},anchor=north},
    y label style={at={(axis description cs:0.2,0.5)},anchor=south},
]
\addplot[name path=companion_upper,draw=none,forget plot] table[x=cover,y expr=\thisrow{companion_acc_avg}+\thisrow{companion_acc_std}, col sep=comma] {\file};
\addplot[name path=companion_lower,draw=none,forget plot] table[x=cover,y expr=\thisrow{companion_acc_avg}-\thisrow{companion_acc_std}, col sep=comma] {\file};
\addplot [fill=blue!30,opacity=0.5,forget plot] fill between[of=companion_upper and companion_lower];

\addplot[name path=corels_upper,draw=none,forget plot] table[x=cover,y expr=\thisrow{corels_acc_avg}+\thisrow{corels_acc_std}, col sep=comma] {\file};
\addplot[name path=corels_lower,draw=none,forget plot] table[x=cover,y expr=\thisrow{corels_acc_avg}-\thisrow{corels_acc_std}, col sep=comma] {\file};
\addplot [fill=red!30,opacity=0.5,forget plot] fill between[of=corels_upper and corels_lower];

\addplot[name path=sbrl_upper,draw=none,forget plot] table[x=cover,y expr=\thisrow{sbrl_acc_avg}+\thisrow{sbrl_acc_std}, col sep=comma] {\file};
\addplot[name path=sbrl_lower,draw=none,forget plot] table[x=cover,y expr=\thisrow{sbrl_acc_avg}-\thisrow{sbrl_acc_std}, col sep=comma] {\file};
\addplot [fill=green!30,opacity=0.5,forget plot] fill between[of=sbrl_upper and sbrl_lower];

\addplot[blue, solid, line width=2pt] table[x=cover,y=companion_acc_avg,col sep=comma] {\file};
\addplot[red, dashed, line width=2pt] table[x=cover,y=corels_acc_avg,col sep=comma] {\file};
\addplot[green!40!black!60!, densely dotted, line width=2pt] table[x=cover,y=sbrl_acc_avg,col sep=comma] {\file};

\addplot[black, fill=magenta!99!white, mark=square*, mark size=4pt] table[x=cover,y=c45_acc_avg,col sep=comma] {\tree};
\addplot[black, fill=cyan!99!white, mark=diamond*, mark size=5pt] table[x=cover,y=cart_acc_avg,col sep=comma] {\tree};

\end{axis}
\node[below=1.1cm of axis.south] {\leg};

\end{tikzpicture}%

    \hspace{-4pt}
  \def\file{./figs/juvenile_xg_avg.csv}%
  \def\tree{./figs/juvenile_tree.csv}%
  \def\ymax{0.91}%
  \def\ymin{0.86}%
  \def\leg{(d) juvenile}%
  \begin{tikzpicture}
\begin{axis}[
    name=axis,
    width=0.24\textwidth,
    xmin=0, xmax=1,
    ymin=\ymin, ymax=\ymax,
    xlabel={\footnotesize Transparency $T_m$},
    ylabel={\footnotesize Accuracy $A_m$},
    tick label style={font=\scriptsize},
    x label style={at={(axis description cs:0.5,0.05)},anchor=north},
    y label style={at={(axis description cs:0.2,0.5)},anchor=south},
]
\addplot[name path=companion_upper,draw=none,forget plot] table[x=cover,y expr=\thisrow{companion_acc_avg}+\thisrow{companion_acc_std}, col sep=comma] {\file};
\addplot[name path=companion_lower,draw=none,forget plot] table[x=cover,y expr=\thisrow{companion_acc_avg}-\thisrow{companion_acc_std}, col sep=comma] {\file};
\addplot [fill=blue!30,opacity=0.5,forget plot] fill between[of=companion_upper and companion_lower];

\addplot[name path=corels_upper,draw=none,forget plot] table[x=cover,y expr=\thisrow{corels_acc_avg}+\thisrow{corels_acc_std}, col sep=comma] {\file};
\addplot[name path=corels_lower,draw=none,forget plot] table[x=cover,y expr=\thisrow{corels_acc_avg}-\thisrow{corels_acc_std}, col sep=comma] {\file};
\addplot [fill=red!30,opacity=0.5,forget plot] fill between[of=corels_upper and corels_lower];

\addplot[name path=sbrl_upper,draw=none,forget plot] table[x=cover,y expr=\thisrow{sbrl_acc_avg}+\thisrow{sbrl_acc_std}, col sep=comma] {\file};
\addplot[name path=sbrl_lower,draw=none,forget plot] table[x=cover,y expr=\thisrow{sbrl_acc_avg}-\thisrow{sbrl_acc_std}, col sep=comma] {\file};
\addplot [fill=green!30,opacity=0.5,forget plot] fill between[of=sbrl_upper and sbrl_lower];

\addplot[blue, solid, line width=2pt] table[x=cover,y=companion_acc_avg,col sep=comma] {\file};
\addplot[red, dashed, line width=2pt] table[x=cover,y=corels_acc_avg,col sep=comma] {\file};
\addplot[green!40!black!60!, densely dotted, line width=2pt] table[x=cover,y=sbrl_acc_avg,col sep=comma] {\file};

\addplot[black, fill=magenta!99!white, mark=square*, mark size=4pt] table[x=cover,y=c45_acc_avg,col sep=comma] {\tree};
\addplot[black, fill=cyan!99!white, mark=diamond*, mark size=5pt] table[x=cover,y=cart_acc_avg,col sep=comma] {\tree};

\end{axis}
\node[below=1.1cm of axis.south] {\leg};

\end{tikzpicture}%
\\
  \def\file{./figs/frisk_xg_avg.csv}%
  \def\tree{./figs/frisk_tree.csv}%
  \def\ymax{0.70}%
  \def\ymin{0.67}%
  \def\leg{(e) frisk}%
  \begin{tikzpicture}
\begin{axis}[
    name=axis,
    width=0.24\textwidth,
    xmin=0, xmax=1,
    ymin=\ymin, ymax=\ymax,
    xlabel={\footnotesize Transparency $T_m$},
    ylabel={\footnotesize Accuracy $A_m$},
    tick label style={font=\scriptsize},
    x label style={at={(axis description cs:0.5,0.05)},anchor=north},
    y label style={at={(axis description cs:0.2,0.5)},anchor=south},
]
\addplot[name path=companion_upper,draw=none,forget plot] table[x=cover,y expr=\thisrow{companion_acc_avg}+\thisrow{companion_acc_std}, col sep=comma] {\file};
\addplot[name path=companion_lower,draw=none,forget plot] table[x=cover,y expr=\thisrow{companion_acc_avg}-\thisrow{companion_acc_std}, col sep=comma] {\file};
\addplot [fill=blue!30,opacity=0.5,forget plot] fill between[of=companion_upper and companion_lower];

\addplot[name path=corels_upper,draw=none,forget plot] table[x=cover,y expr=\thisrow{corels_acc_avg}+\thisrow{corels_acc_std}, col sep=comma] {\file};
\addplot[name path=corels_lower,draw=none,forget plot] table[x=cover,y expr=\thisrow{corels_acc_avg}-\thisrow{corels_acc_std}, col sep=comma] {\file};
\addplot [fill=red!30,opacity=0.5,forget plot] fill between[of=corels_upper and corels_lower];

\addplot[name path=sbrl_upper,draw=none,forget plot] table[x=cover,y expr=\thisrow{sbrl_acc_avg}+\thisrow{sbrl_acc_std}, col sep=comma] {\file};
\addplot[name path=sbrl_lower,draw=none,forget plot] table[x=cover,y expr=\thisrow{sbrl_acc_avg}-\thisrow{sbrl_acc_std}, col sep=comma] {\file};
\addplot [fill=green!30,opacity=0.5,forget plot] fill between[of=sbrl_upper and sbrl_lower];

\addplot[blue, solid, line width=2pt] table[x=cover,y=companion_acc_avg,col sep=comma] {\file};
\addplot[red, dashed, line width=2pt] table[x=cover,y=corels_acc_avg,col sep=comma] {\file};
\addplot[green!40!black!60!, densely dotted, line width=2pt] table[x=cover,y=sbrl_acc_avg,col sep=comma] {\file};

\addplot[black, fill=magenta!99!white, mark=square*, mark size=4pt] table[x=cover,y=c45_acc_avg,col sep=comma] {\tree};
\addplot[black, fill=cyan!99!white, mark=diamond*, mark size=5pt] table[x=cover,y=cart_acc_avg,col sep=comma] {\tree};

\end{axis}
\node[below=1.1cm of axis.south] {\leg};

\end{tikzpicture}%

    \hspace{-4pt}
  \def\file{./figs/recidivism_xg_avg.csv}%
  \def\tree{./figs/recidivism_tree.csv}%
  \def\ymax{0.78}%
  \def\ymin{0.70}%
  \def\leg{(f) recidivism}%
  \begin{tikzpicture}
\begin{axis}[
    name=axis,
    width=0.24\textwidth,
    xmin=0, xmax=1,
    ymin=\ymin, ymax=\ymax,
    xlabel={\footnotesize Transparency $T_m$},
    ylabel={\footnotesize Accuracy $A_m$},
    tick label style={font=\scriptsize},
    x label style={at={(axis description cs:0.5,0.05)},anchor=north},
    y label style={at={(axis description cs:0.2,0.5)},anchor=south},
]
\addplot[name path=companion_upper,draw=none,forget plot] table[x=cover,y expr=\thisrow{companion_acc_avg}+\thisrow{companion_acc_std}, col sep=comma] {\file};
\addplot[name path=companion_lower,draw=none,forget plot] table[x=cover,y expr=\thisrow{companion_acc_avg}-\thisrow{companion_acc_std}, col sep=comma] {\file};
\addplot [fill=blue!30,opacity=0.5,forget plot] fill between[of=companion_upper and companion_lower];

\addplot[name path=corels_upper,draw=none,forget plot] table[x=cover,y expr=\thisrow{corels_acc_avg}+\thisrow{corels_acc_std}, col sep=comma] {\file};
\addplot[name path=corels_lower,draw=none,forget plot] table[x=cover,y expr=\thisrow{corels_acc_avg}-\thisrow{corels_acc_std}, col sep=comma] {\file};
\addplot [fill=red!30,opacity=0.5,forget plot] fill between[of=corels_upper and corels_lower];

\addplot[name path=sbrl_upper,draw=none,forget plot] table[x=cover,y expr=\thisrow{sbrl_acc_avg}+\thisrow{sbrl_acc_std}, col sep=comma] {\file};
\addplot[name path=sbrl_lower,draw=none,forget plot] table[x=cover,y expr=\thisrow{sbrl_acc_avg}-\thisrow{sbrl_acc_std}, col sep=comma] {\file};
\addplot [fill=green!30,opacity=0.5,forget plot] fill between[of=sbrl_upper and sbrl_lower];

\addplot[blue, solid, line width=2pt] table[x=cover,y=companion_acc_avg,col sep=comma] {\file};
\addplot[red, dashed, line width=2pt] table[x=cover,y=corels_acc_avg,col sep=comma] {\file};
\addplot[green!40!black!60!, densely dotted, line width=2pt] table[x=cover,y=sbrl_acc_avg,col sep=comma] {\file};

\addplot[black, fill=magenta!99!white, mark=square*, mark size=4pt] table[x=cover,y=c45_acc_avg,col sep=comma] {\tree};
\addplot[black, fill=cyan!99!white, mark=diamond*, mark size=5pt] table[x=cover,y=cart_acc_avg,col sep=comma] {\tree};

\end{axis}
\node[below=1.1cm of axis.south] {\leg};

\end{tikzpicture}%

    \hspace{-4pt}
  \def\file{./figs/magic_xg_avg.csv}%
  \def\tree{./figs/magic_tree.csv}%
  \def\ymax{0.89}%
  \def\ymin{0.80}%
  \def\leg{(g) magic}%
  \begin{tikzpicture}
\begin{axis}[
    name=axis,
    width=0.24\textwidth,
    xmin=0, xmax=1,
    ymin=\ymin, ymax=\ymax,
    xlabel={\footnotesize Transparency $T_m$},
    ylabel={\footnotesize Accuracy $A_m$},
    tick label style={font=\scriptsize},
    x label style={at={(axis description cs:0.5,0.05)},anchor=north},
    y label style={at={(axis description cs:0.2,0.5)},anchor=south},
]
\addplot[name path=companion_upper,draw=none,forget plot] table[x=cover,y expr=\thisrow{companion_acc_avg}+\thisrow{companion_acc_std}, col sep=comma] {\file};
\addplot[name path=companion_lower,draw=none,forget plot] table[x=cover,y expr=\thisrow{companion_acc_avg}-\thisrow{companion_acc_std}, col sep=comma] {\file};
\addplot [fill=blue!30,opacity=0.5,forget plot] fill between[of=companion_upper and companion_lower];

\addplot[name path=corels_upper,draw=none,forget plot] table[x=cover,y expr=\thisrow{corels_acc_avg}+\thisrow{corels_acc_std}, col sep=comma] {\file};
\addplot[name path=corels_lower,draw=none,forget plot] table[x=cover,y expr=\thisrow{corels_acc_avg}-\thisrow{corels_acc_std}, col sep=comma] {\file};
\addplot [fill=red!30,opacity=0.5,forget plot] fill between[of=corels_upper and corels_lower];

\addplot[name path=sbrl_upper,draw=none,forget plot] table[x=cover,y expr=\thisrow{sbrl_acc_avg}+\thisrow{sbrl_acc_std}, col sep=comma] {\file};
\addplot[name path=sbrl_lower,draw=none,forget plot] table[x=cover,y expr=\thisrow{sbrl_acc_avg}-\thisrow{sbrl_acc_std}, col sep=comma] {\file};
\addplot [fill=green!30,opacity=0.5,forget plot] fill between[of=sbrl_upper and sbrl_lower];

\addplot[blue, solid, line width=2pt] table[x=cover,y=companion_acc_avg,col sep=comma] {\file};
\addplot[red, dashed, line width=2pt] table[x=cover,y=corels_acc_avg,col sep=comma] {\file};
\addplot[green!40!black!60!, densely dotted, line width=2pt] table[x=cover,y=sbrl_acc_avg,col sep=comma] {\file};

\addplot[black, fill=magenta!99!white, mark=square*, mark size=4pt] table[x=cover,y=c45_acc_avg,col sep=comma] {\tree};
\addplot[black, fill=cyan!99!white, mark=diamond*, mark size=5pt] table[x=cover,y=cart_acc_avg,col sep=comma] {\tree};

\end{axis}
\node[below=1.1cm of axis.south] {\leg};

\end{tikzpicture}%

    \hspace{-4pt}
  \def\file{./figs/coupon_xg_avg.csv}%
  \def\tree{./figs/coupon_tree.csv}%
  \def\ymax{0.80}%
  \def\ymin{0.65}%
  \def\leg{(h) coupon}%
  \begin{tikzpicture}
\begin{axis}[
    name=axis,
    width=0.24\textwidth,
    xmin=0, xmax=1,
    ymin=\ymin, ymax=\ymax,
    xlabel={\footnotesize Transparency $T_m$},
    ylabel={\footnotesize Accuracy $A_m$},
    tick label style={font=\scriptsize},
    x label style={at={(axis description cs:0.5,0.05)},anchor=north},
    y label style={at={(axis description cs:0.2,0.5)},anchor=south},
]
\addplot[name path=companion_upper,draw=none,forget plot] table[x=cover,y expr=\thisrow{companion_acc_avg}+\thisrow{companion_acc_std}, col sep=comma] {\file};
\addplot[name path=companion_lower,draw=none,forget plot] table[x=cover,y expr=\thisrow{companion_acc_avg}-\thisrow{companion_acc_std}, col sep=comma] {\file};
\addplot [fill=blue!30,opacity=0.5,forget plot] fill between[of=companion_upper and companion_lower];

\addplot[name path=corels_upper,draw=none,forget plot] table[x=cover,y expr=\thisrow{corels_acc_avg}+\thisrow{corels_acc_std}, col sep=comma] {\file};
\addplot[name path=corels_lower,draw=none,forget plot] table[x=cover,y expr=\thisrow{corels_acc_avg}-\thisrow{corels_acc_std}, col sep=comma] {\file};
\addplot [fill=red!30,opacity=0.5,forget plot] fill between[of=corels_upper and corels_lower];

\addplot[name path=sbrl_upper,draw=none,forget plot] table[x=cover,y expr=\thisrow{sbrl_acc_avg}+\thisrow{sbrl_acc_std}, col sep=comma] {\file};
\addplot[name path=sbrl_lower,draw=none,forget plot] table[x=cover,y expr=\thisrow{sbrl_acc_avg}-\thisrow{sbrl_acc_std}, col sep=comma] {\file};
\addplot [fill=green!30,opacity=0.5,forget plot] fill between[of=sbrl_upper and sbrl_lower];

\addplot[blue, solid, line width=2pt] table[x=cover,y=companion_acc_avg,col sep=comma] {\file};
\addplot[red, dashed, line width=2pt] table[x=cover,y=corels_acc_avg,col sep=comma] {\file};
\addplot[green!40!black!60!, densely dotted, line width=2pt] table[x=cover,y=sbrl_acc_avg,col sep=comma] {\file};

\addplot[black, fill=magenta!99!white, mark=square*, mark size=4pt] table[x=cover,y=c45_acc_avg,col sep=comma] {\tree};
\addplot[black, fill=cyan!99!white, mark=diamond*, mark size=5pt] table[x=cover,y=cart_acc_avg,col sep=comma] {\tree};

\end{axis}
\node[below=1.1cm of axis.south] {\leg};

\end{tikzpicture}%

    \caption{The transparency--accuracy trade-off (with XGBoost as $f_b$): The solid lines denote average trade-off curves, while the shaded regions denote $\pm$ standard deviations evaluated via 5-fold cross validation.}
    \label{fig:xg}
\end{figure*}

\begin{table}[!h]
\caption{AUTACs on AdaBoost: The numbers in the parenthesis denote standard deviations. Bold fonts denote the best results (underlined), and the results which was not significantly different from the best result (t-test with the 5\% significance level).}
\label{tab:auc_ada}
\small
\begin{tabular}{llll}
\toprule
{} &                             CRL &             CORELS    &             SBRL         \\
\midrule
messidor   &  \underline{\textbf{.675 (.010)}} &  \textbf{.674 (.013)} &  \textbf{.660 (.020)} \\
german     &  \underline{\textbf{.754 (.010)}} &           .738 (.013) &  \textbf{.749 (.005)} \\
adult      &  \underline{\textbf{.856 (.002)}} &           .837 (.004) &           .850 (.004) \\
juvenile   &  \underline{\textbf{.892 (.005)}} &  \textbf{.885 (.013)} &           .883 (.005) \\
frisk      &  \underline{\textbf{.685 (.003)}} &           .682 (.003) &           .678 (.002) \\
recidivism &  \underline{\textbf{.763 (.006)}} &           .744 (.007) &           .759 (.006) \\
magic      &  \underline{\textbf{.864 (.007)}} &           .841 (.010) &           .852 (.007) \\
coupon     &  \textbf{.740 (.012)} &           .714 (.015) &  \underline{\textbf{.743 (.017)}} \\
\bottomrule
\end{tabular}
\end{table}\normalsize

\begin{table}[!h]
\caption{AUTACs on XGBoost: The numbers in the parenthesis denote standard deviations. Bold fonts denote the best results (underlined), and the results which was not significantly different from the best result (t-test with the 5\% significance level).}
\label{tab:auc_xg}
\small
\begin{tabular}{llll}
\toprule
{} &                             CRL &             CORELS    &             SBRL         \\
\midrule
messidor   &  \underline{\textbf{.689 (.017)}} &  \textbf{.681 (.019)} &  \textbf{.670 (.022)} \\
german     &  \underline{\textbf{.748 (.022)}} &  \textbf{.732 (.016)} &  \textbf{.734 (.010)} \\
adult      &  \underline{\textbf{.856 (.002)}} &           .837 (.004) &           .851 (.004) \\
juvenile   &  \underline{\textbf{.898 (.006)}} &  \textbf{.888 (.015)} &           .884 (.005) \\
frisk      &  \underline{\textbf{.686 (.003)}} &           .683 (.003) &           .679 (.003) \\
recidivism &  \underline{\textbf{.766 (.007)}} &           .744 (.008) &           .759 (.006) \\
magic      &  \underline{\textbf{.863 (.004)}} &           .840 (.007) &           .853 (.004) \\
coupon     &  \textbf{.739 (.013)} &           .713 (.015) &  \underline{\textbf{.744 (.017)}} \\
\bottomrule
\end{tabular}
\end{table}\normalsize

%%%%%%%%%%%%%%%%%%%%%%%%%%%%%%%%%%%%%%%%%%%%%%%%%%%%%%%%%%%%%%%%%%%%%%%%
\clearpage
\section{Survey}
\label{app:survey}

\subsection{Survey Question}

\begin{figure*}[h]
    \centering
    \includegraphics[width=0.8\linewidth]{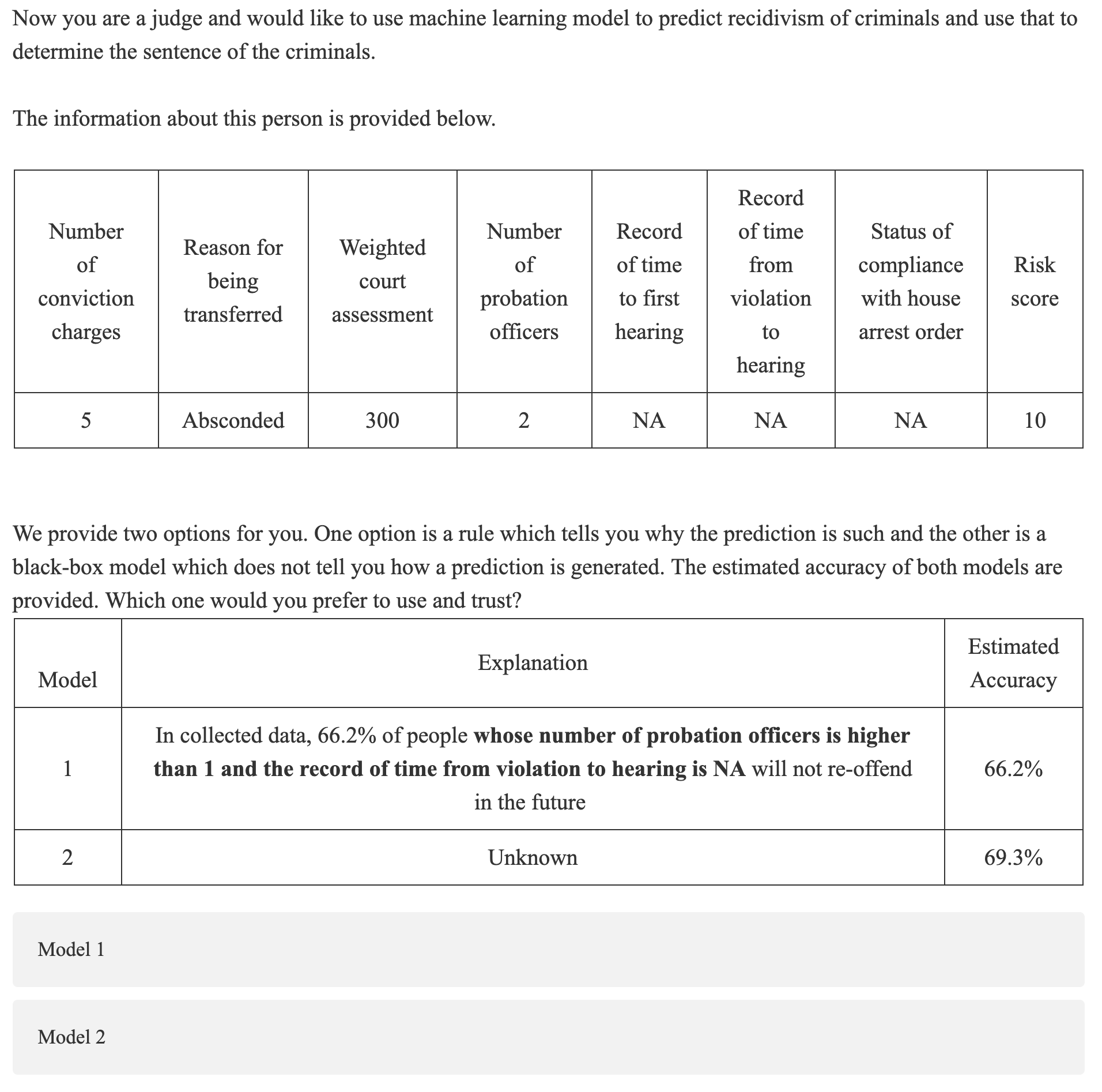}
    \caption{An example of our survey questions.}
    \label{fig:survey} 
\end{figure*}

\figurename~\ref{fig:survey} is an example of our survey question.

\subsection{Additional Results}

We analyzed the trade-off curve in different groups of users, by gender and by age, and summarized the results in Figures~\ref{fig:trade-off-gender} and \ref{fig:trade-off-age}.
The results suggest that gender does not seem to play any role in the preference for interpretability but the younger group (30 years old or younger) are more likely to choose rules over black-box models than the older group (older than 30 years old). These results suggest that the users indeed have their own preference on the trade-off between the accuracy and transparency. Thus, it is essential to provide the users a freedom of choosing the trade-off depending on their preference.

\begin{figure}[h]
    \centering
    \begin{tikzpicture}
    \begin{axis}[
        width=0.4\textwidth,
        height=0.27\textwidth,
        xmin=-0.25, xmax=0.05,
        ymin=0.2, ymax=1.0,
        xlabel={\footnotesize Relative accuracy loss if using a rule},
        ylabel style={align=center},
        ylabel={\footnotesize Ratios of the participants\\ preferred a rule},
        xtick={-0.25,-0.2,-0.15,-0.10,-0.05,0.0,0.05},
        xticklabel style={
        	/pgf/number format/fixed,
            /pgf/number format/precision=2,
            /pgf/number format/fixed zerofill
        },
        legend pos=north west,
        tick label style={font=\scriptsize},
        scatter/classes={a={mark=o,draw=blue,fill=blue}},
    ]
    \addplot[only marks,mark=*,draw=blue,fill=blue]
    table[meta=label] {
    x y label
-0.008290155440414507 0.8928571428571429 a
-0.012282497441146366 0.7419354838709677 a
-0.0735981308411215 0.3 a
-0.041343669250646 0.7586206896551724 a
-0.029126213592233007 0.75 a
-0.08502024291497975 0.56 a
-0.2402496099843994 0.24 a
0.0339943342776204 1.0 a
0.006544502617801046 1.0 a
-0.0059031877213695395 0.8 a
-0.044733044733044736 0.5862068965517241 a
-0.14027777777777778 0.4827586206896552 a
-0.2730109204368175 0.25806451612903225 a
0.03443877551020408 1.0 a
-0.051696284329563816 0.5333333333333333 a
-0.026683608640406607 0.75 a
0.0 0.9666666666666667 a
-0.06424242424242424 0.5666666666666667 a
    };
    \addplot[only marks,mark=square*,draw=red,fill=red]
    table[meta=label] {
    x y label
-0.008290155440414507 0.7857142857142857 b
-0.012282497441146366 0.9285714285714286 b
-0.0735981308411215 0.5454545454545454 b
-0.041343669250646 0.75 b
-0.029126213592233007 0.8461538461538461 b
-0.08502024291497975 0.4166666666666667 b
-0.2402496099843994 0.3333333333333333 b
0.0339943342776204 1.0 b
0.006544502617801046 1.0 b
-0.0059031877213695395 0.9411764705882353 b
-0.044733044733044736 0.6 b
-0.14027777777777778 0.6 b
-0.2730109204368175 0.25 b
0.03443877551020408 1.0 b
-0.051696284329563816 0.7692307692307693 b
-0.026683608640406607 0.6923076923076923 b
0.0 1.0 b
-0.06424242424242424 0.5 b
    };
    \legend{Male, Female}
    \end{axis}
    \end{tikzpicture}
    \caption{Human evaluated trade-off between model transparency and accuracy: Male vs.\ Female}
    \label{fig:trade-off-gender} 
\end{figure}

\begin{figure}[h]
    \centering
    \begin{tikzpicture}
    \begin{axis}[
        width=0.4\textwidth,
        height=0.27\textwidth,
        xmin=-0.25, xmax=0.05,
        ymin=0.2, ymax=1.0,
        xlabel={\footnotesize Relative accuracy loss if using a rule},
        ylabel style={align=center},
        ylabel={\footnotesize Ratios of the participants\\ preferred a rule},
        xtick={-0.25,-0.2,-0.15,-0.10,-0.05,0.0,0.05},
        xticklabel style={
        	/pgf/number format/fixed,
            /pgf/number format/precision=2,
            /pgf/number format/fixed zerofill
        },
        legend pos=north west,
        tick label style={font=\scriptsize},
        scatter/classes={a={mark=o,draw=blue,fill=blue}},
    ]
    \addplot[only marks,mark=*,draw=blue,fill=blue]
    table[meta=label] {
    x y label
-0.008290155440414507 0.90625 a
-0.012282497441146366 0.8484848484848485 a
-0.0735981308411215 0.35135135135135137 a
-0.041343669250646 0.75 a
-0.029126213592233007 0.8108108108108109 a
-0.08502024291497975 0.4444444444444444 a
-0.2402496099843994 0.3055555555555556 a
0.0339943342776204 0.9682539682539683 a
0.006544502617801046 0.9411764705882353 a
-0.0059031877213695395 0.85 a
-0.044733044733044736 0.6486486486486487 a
-0.14027777777777778 0.525 a
-0.2730109204368175 0.2857142857142857 a
0.03443877551020408 0.96875 a
-0.051696284329563816 0.5945945945945946 a
-0.026683608640406607 0.7368421052631579 a
0.0 0.9166666666666666 a
-0.06424242424242424 0.5675675675675675 a
    };
    \addplot[only marks,mark=square*,draw=red,fill=red]
    table[meta=label] {
    x y label
-0.008290155440414507 0.6923076923076923 b
-0.012282497441146366 0.5333333333333333 b
-0.0735981308411215 0.4 b
-0.041343669250646 0.5555555555555556 b
-0.029126213592233007 0.5555555555555556 b
-0.08502024291497975 0.4444444444444444 b
-0.2402496099843994 0.18181818181818182 b
0.0339943342776204 0.8095238095238095 b
0.006544502617801046 0.8571428571428571 b
-0.0059031877213695395 0.6363636363636364 b
-0.044733044733044736 0.5833333333333334 b
-0.14027777777777778 0.375 b
-0.2730109204368175 0.23076923076923078 b
0.03443877551020408 0.7619047619047619 b
-0.051696284329563816 0.5833333333333334 b
-0.026683608640406607 0.7 b
0.0 0.8666666666666667 b
-0.06424242424242424 0.45454545454545453 b
    };
    \legend{Young, Old}
    \end{axis}
    \end{tikzpicture}
    \caption{Human evaluated trade-off between model transparency and accuracy: Young vs.\ Old}
    \label{fig:trade-off-age} 
\end{figure}

\end{document}